\documentclass[12pt]{article}
\usepackage{scicite}
\usepackage{times}
\usepackage{amssymb,amsfonts,amsmath}
\usepackage{amsthm}
\newtheorem{theorem}{Theorem}[section]
\newtheorem{lemma}[theorem]{Lemma}
\newtheorem{proposition}[theorem]{Proposition}

\newtheorem{definition}[theorem]{Definition}
\newtheorem{remark}[theorem]{Remark}

\usepackage{booktabs}
\usepackage{algorithm}
\usepackage[end]{algpseudocode}

\usepackage{graphicx}
\usepackage[font=small,labelfont=bf,labelsep=period]{caption}
\usepackage{todonotes}
\usepackage{stmaryrd}
\usepackage{comment}
\newcommand{\R}[0] {{ \mathbb{R} }}

\newcommand{\Expect}[0] {{ \mathbb{E} }}
\newcommand{\MaxExp}[0] {{ \mathbb{M} }}

\DeclareMathOperator*{\argmax}{argmax}
\newcommand{\D}[0] {{ \mathcal{D} }}
\newcommand{\dg}[0] {{ \dagger }}
\newcommand{\Cl}[0] {{ \mathcal{C} }}
\newcommand{\G}[0] {{ \mathcal{G} }}
\newcommand{\HH}[0] {{ \mathcal{H} }}
\newcommand{\A}[0] {{ \mathcal{A} }}
\newcommand{\Pa}[0] {{ \mathcal{P} }}
\newcommand{\W}[0] {{ \mathcal{W} }}
\newcommand{\M}[0] {{ \mathcal{M} }}
\newcommand{\T}[0] {{ \mathcal{T} }}
\newcommand{\ith}[0] {{ \emph{i}-theory }}
\newcommand{\drelax}[0] {{ \,\,\rightarrow_{d}\,\, }}
\newcommand{\lcmatch}[0] {{ \,\,\star_{LC}\,\, }}
\newcommand{\cnnconv}[0] {{ \,\,\star_{\textrm{DCN}}\,\, }}

\DeclareMathOperator*{\MaxPool}{MaxPool}

\usepackage{color}

\newenvironment{sciabstract}{%
\begin{quote} \bf}
{\end{quote}}

\newcounter{lastnote}

\title{\bf A Probabilistic Theory of Deep Learning}

\author
{Ankit B.\ Patel, 
Tan Nguyen, Richard G.\ Baraniuk \\[4mm]
\normalsize{Department of Electrical and Computer Engineering} \\
\normalsize{Rice University} \\
\normalsize{\{abp4, mn15, richb\}@rice.edu} \\
\normalsize{April 2, 2015}
}

\date{}

\begin{document} 

\baselineskip16pt

\maketitle 

\begin{sciabstract}

A grand challenge in machine learning is the development of computational algorithms that match or outperform humans in perceptual inference tasks such as visual object and speech recognition.  
The key factor complicating such tasks is the presence of numerous {\em nuisance variables}, for instance, the unknown object position, orientation, and scale in object recognition or the unknown voice pronunciation, pitch, and speed in speech recognition.
Recently, a new breed of {\em deep learning} algorithms have emerged for high-nuisance inference tasks; they are constructed from many layers of alternating linear and nonlinear processing units and are trained using large-scale algorithms and massive amounts of training data. 
The recent success of deep learning systems is impressive --- they now routinely yield pattern recognition systems with near- or super-human capabilities --- but a fundamental question remains: {\em Why do they work?}
Intuitions abound, but a coherent framework for understanding, analyzing, and synthesizing deep learning architectures has remained elusive.

We answer this question by developing a new probabilistic framework for deep learning based on a Bayesian {\em generative probabilistic model} that explicitly captures variation due to nuisance variables.  
The graphical structure of the model enables it to be learned from data using classical expectation-maximization techniques.
Furthermore, by relaxing the generative model to a discriminative one, we can recover two of the
current leading deep learning systems, deep convolutional neural networks (DCNs) and random decision forests (RDFs),
providing insights into their successes and shortcomings as well as a principled route to their improvement. 
\end{sciabstract}

\newpage
\tableofcontents
\newpage

\section{Introduction}

Humans are expert at a wide array of complicated sensory inference tasks, from recognizing objects in an image to understanding phonemes in a speech signal, despite significant variations such as the position, orientation, and scale of objects and the pronunciation, pitch, and volume of speech.
Indeed, the main challenge in many sensory perception tasks in vision, speech, and natural language processing is a high amount of such {\em nuisance variation}. 
Nuisance variations complicate perception, because they turn otherwise simple statistical inference problems with a small number of variables (e.g., class label) into much higher-dimensional problems.
For example, images of a car taken from different camera viewpoints lie on a highly curved, nonlinear manifold in high-dimensional space that is intertwined with the manifolds of myriad other objects.  
The key challenge in developing an inference algorithm is then \textit{how to factor out all of the nuisance variation in the input}. 
Over the past few decades, a vast literature that approaches this problem from myriad different perspectives has developed, but the most difficult inference problems have remained out of reach.

Recently, a new breed of machine learning algorithms have emerged for high-nuisance inference tasks, resulting in pattern recognition systems with sometimes super-human capabilities \cite{schmidhuber2015deep}.  
These so-called {\em deep learning} systems share two common hallmarks. 
First, architecturally, they are constructed from many layers of alternating linear and nonlinear processing units.
Second, computationally, their parameters are learned using large-scale algorithms and massive amounts of training data. 
Two examples of such architectures are the {\em deep convolutional neural network} (DCN), which has seen great success in tasks like visual object recognition and localization \cite{zeiler2014visualizing}, speech recognition \cite{hannun2014deepspeech}, and part-of-speech recognition \cite{Schmid:1994:PTN:991886.991915}, and  {\em random decision forests (RDFs)} \cite{criminisi2013decision} for image segmentation. 
The success of deep learning systems is impressive, but a fundamental question remains: {\em Why do they work?} 
Intuitions abound to explain their success. Some explanations focus on properties of feature invariance and selectivity developed over multiple layers, while others credit raw computational power and the amount of available training data \cite{schmidhuber2015deep}. 
However, beyond these intuitions, a coherent theoretical framework for understanding, analyzing, and synthesizing deep learning architectures has remained elusive. 

In this paper, we develop a new theoretical framework that provides insights into both the successes and shortcomings of deep learning systems, as well as a principled route to their design and improvement.
Our framework is based on a {\em generative probabilistic model that explicitly captures variation due to latent nuisance variables}.  
The {\em Rendering Model} (RM) explicitly models nuisance variation through a {\em rendering function} that combines the task-specific variables of interest (e.g., object class in an object recognition task) and the collection of nuisance variables.  
The {\em Deep Rendering Model} (DRM) extends the RM in a hierarchical fashion by rendering via a product of affine nuisance transformations across multiple levels of abstraction.  
The graphical structures of the RM and DRM enable inference via message passing, using, for example, the sum-product or max-sum algorithms, and training via the expectation-maximization (EM) algorithm.
A key element of the framework is the relaxation of the RM/DRM generative model to a discriminative one in order to optimize the bias-variance tradeoff.

The DRM unites and subsumes two of the current leading deep learning based systems as {\em max-sum message passing networks}.   
That is, configuring the DRM with two different nuisance structures --- Gaussian translational nuisance or 
evolutionary additive nuisance --- leads directly to DCNs and RDFs, respectively.
The intimate connection between the DRM and these deep learning systems provides a range of new insights into how and why they work, answering several open questions.  
Moreover, the DRM provides insights into how and why deep learning fails and suggests pathways to their improvement.

It is important to note that our theory and methods apply to a wide range of different inference tasks (including, for example, classification, estimation, regression, etc.) 
that feature a number of task-irrelevant nuisance variables
(including, for example, object and speech recognition). 
However, for concreteness of exposition, we will focus below on the classification problem underlying visual object recognition. 

This paper is organized as follows.  
Section \ref{sec:drm} introduces the RM and DRM and demonstrates step-by-step how they map onto DCNs.
Section \ref{sec:insights} then summarizes some of the key insights that the DRM provides into the operation and performance of DCNs.
Section \ref{sec:rdf} 
proceeds in a similar fashion to derive RDFs from a variant of the DRM that models a hierarchy of categories.
Section \ref{sec:ext} closes the paper by suggesting a number of promising avenues for research, including several that should lead to improvement in deep learning system performance and generality.  
The proofs of several results appear in the Appendix.

\section{A Deep Probabilistic Model for Nuisance Variation}
\label{sec:drm}

This section develops the RM, a generative probabilistic model that \textit{explicitly captures nuisance transformations as latent variables}. 
We show how inference in the RM corresponds to operations in a single layer of a DCN. 
We then extend the RM by defining the DRM, a rendering model with layers representing different scales or levels of abstraction. 
Finally, we show that, after the application of a discriminative relaxation, inference and learning in the DRM correspond to feedforward propagation and back propagation training in the DCN. This enables us to conclude that DCNs are probabilistic message passing networks, thus unifying the probabilistic and neural network perspectives.

\subsection{The Rendering Model: Capturing Nuisance Variation}
\label{sec:rm}

Visual object recognition is naturally formulated as a statistical classification problem.\footnote{Recall that we focus on object recognition from images only for concreteness of exposition.}
We are given a $D$-pixel, multi-channel image $I$ of an object, with intensity $I(x, \omega)$ at pixel $x$ and channel $\omega$ (e.g., $\omega= $\{red, green, blue\}).
We seek to infer the object's identity (class) $c \in \Cl$, where $\Cl$ is a finite set of classes.\footnote{The restriction for $\Cl$ to be finite can be removed by using a nonparametric prior such as a Chinese Restaurant Process (CRP) \cite{griffiths2004hierarchical}}
We will use the terms ``object'' and ``class'' interchangeably.
Given a joint probabilistic model $p(I,c)$ for images and objects, we can classify a particular image $I$ using the {\em maximum a posteriori} (MAP) classifier 
\begin{align}
    \label{eqn:map_est_class}
     \hat{c}(I) = \argmax_{c \in \Cl}\, p(c|I) = \argmax_{c \in \Cl}\, p(I|c)p(c),
\end{align}
where $p(I|c)$ is the image likelihood, $p(c)$ is the prior distribution over the classes, and $p(c|I) \propto p(I|c) p(c)$ by Bayes' rule.

Object recognition, like many other inference tasks, is complicated by a high amount of variation due to nuisance variables, which the above formation ignores.  We advocate explicitly modeling nuisance variables by encapsulating all of them into a (possibly high-dimensional) parameter $g \in \G$, where $\G$ is the set of all nuisances. In some cases, it is natural for $g$ to be a transformation and for $\G$ to be endowed with a (semi-)group structure.

We now propose a {\em generative model} for images that explicitly models the relationship between images $I$ of the same object $c$ subject to nuisance $g$.
First, given $c$, $g$, and other auxiliary parameters, we define the {\em rendering function} $R(c,g)$ that renders (produces) an image.
In image inference problems, for example, $R(c,g)$ might be a photorealistic computer graphics engine (c.f., Pixar).
A particular realization of an image is then generated by adding noise to the output of the renderer:
\begin{equation}
    I |c,g = R(c,g) + \textrm{noise}. 
\label{eq:render}
\end{equation}

We assume that the noise distribution is from the exponential family, which includes a large number of practically useful distributions (e.g., Gaussian, Poisson). Also we assume that the noise is independent and identically distributed (iid) as a function of pixel location $x$ and that the class and nuisance variables are independently distributed according to categorical distributions.\footnote{Independence is merely a convenient approximation; in practice, $g$ can depend on $c$.  For example, humans have difficulty recognizing and discriminating upside down faces \cite{Searcy:1996vt}.}
With these assumptions, Eq.~\ref{eq:render} then becomes the probabilistic (shallow) {\em Rendering Model} (RM) 
\begin{align} 
	c &\sim{\rm Cat}(\{\pi_c\}_{c\in\Cl}), \quad\quad \nonumber\\
	g &\sim{\rm Cat}(\{\pi_g\}_{g\in\G}), \quad\quad \nonumber\\	
	I |c,g &\sim \mathcal{Q}(\theta_{cg}).
\label{eqn:RM}
\end{align}
Here $\mathcal{Q}(\theta_{cg})$ denotes a distribution from the exponential family with parameters $\theta_{cg}$, which include the mixing probabilities $\pi_{cg}$, natural parameters $\eta(\theta_{cg})$, sufficient statistics $T(I)$ and whose mean is the rendered template $\mu_{cg}=R(c,g)$.  

\begin{figure}[t!]
   \centering
   \includegraphics[width=0.50\textwidth]{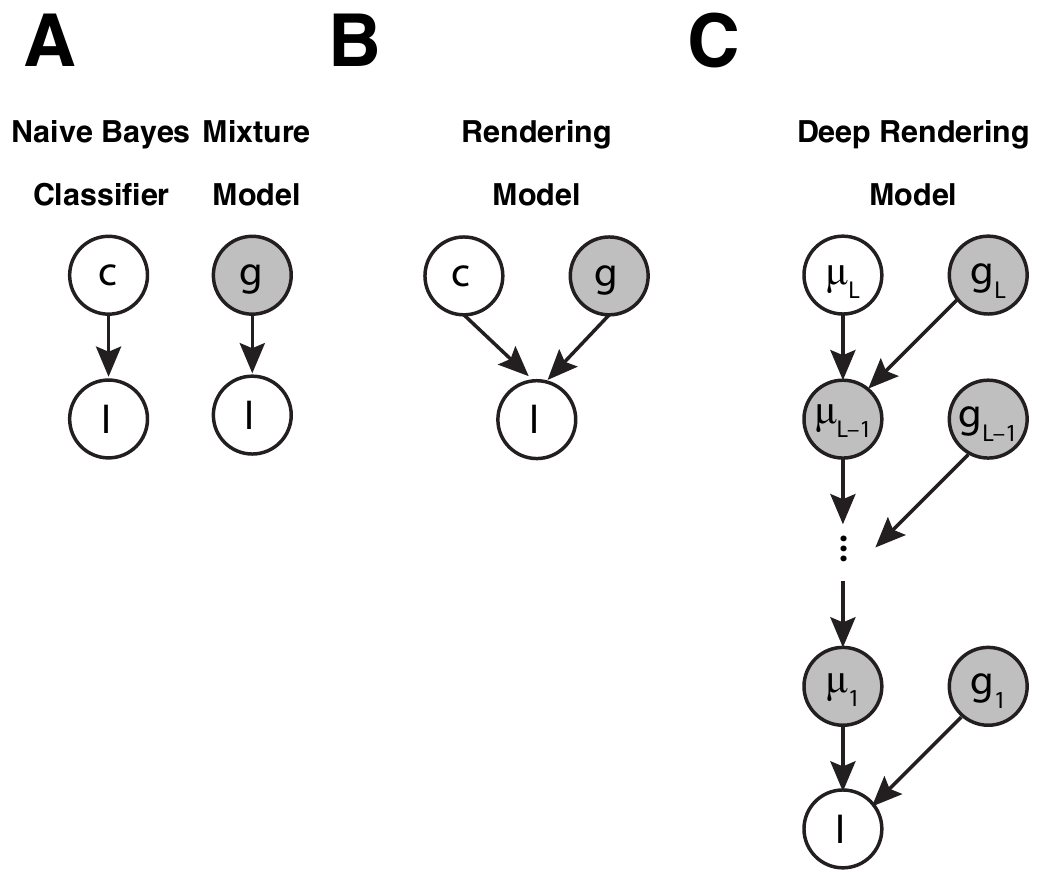} 
   \caption{Graphical depiction of the Naive Bayes Classifier (A, left),  Gaussian Mixture Model (A, right), the shallow Rendering Model (B) and the Deep Rendering Model (C). All dependence on pixel location $x$ has been suppressed for clarity.}
   \label{fig:bigFig}
\end{figure}

An important special case is when $\mathcal{Q}(\theta_{cg})$ is Gaussian, and this defines the {\it Gaussian Rendering Model} (GRM), in which images are generated according to
\begin{align} 
    I | c,g \sim \mathcal{N}(I | \mu_{cg}=R(c,g), \Sigma_{cg}=\sigma^2 \bf{1}),
\end{align}
where $\bf{1}$ is the identity matrix. 
The GRM generalizes both the Gaussian Na\"{i}ve Bayes Classifier (GNBC) and the Gaussian Mixture Model (GMM) by allowing variation in the image to depend on an observed class label $c$, like a GNBC, and on an unobserved nuisance label $g$, like a GMM. The GNBC, GMM and the (G)RM can all be conveniently described as a {\em directed graphical model} \cite{jordan2001graphical}. Figure \ref{fig:bigFig}A depicts the graphical models for the GNBC and GMM, while Fig.\ 1B shows how they are combined to form the (G)RM. 

Finally, since the world is spatially varying and an image can contain a number of different objects, it is natural to break the image up into a number of (overlapping) subimages, called \emph{patches}, that are indexed by spatial location $x$. Thus, a patch is defined here as a collection of pixels centered on a single pixel $x$. In general, patches can overlap, meaning that (i) they do not tile the image, and (ii) an image pixel $x$ can belong to more than one patch. Given this notion of pixels and patches, we allow the class and nuisance variables to depend on pixel/patch location: i.e., local image class $c(x)$ and local nuisance $g(x)$ (see Fig.~\ref{fig:DRM-to-DCN}A).
We will omit the dependence on $x$ when it is clear from context.

The notion of a rendering operator is quite general and can refer to any function that maps a target variable $c$ and nuisance variables $g$ into a pattern or template $R(c,g)$.
For example, in speech recognition, $c$ might be a phoneme, in which case $g$ represents volume, pitch, speed, and accent, and $R(c,g)$ is the amplitude of the acoustic signal (or alternatively the time-frequency representation). 
In natural language processing, $c$ might be the grammatical part-of-speech, in which case $g$ represents syntax and grammar, and $R(c,g)$ is a clause, phrase or sentence.

To perform object recognition with the RM via Eq.~\ref{eqn:map_est_class}, we must marginalize out the nuisance variables $g$. We consider two approaches for doing so, one conventional and one unconventional.
The {\em Sum-Product RM Classifier} (SP-RMC) {\em sums} over all nuisance variables $g \in \G$ and then chooses the most likely class:
\begin{align} 
	\label{eqn:spc}
	\hat{c}_{\rm SP}(I) &= \argmax_{c \in \Cl}  \frac{1}{|\G|} \sum_{g \in \G} p(I|c,g) p(c) p(g) \nonumber \\
	              &= \argmax_{c \in \Cl} \frac{1}{|\G|} \sum_{g \in \G} \exp \langle \eta(\theta_{cg}) | T(I) \rangle,
\end{align}
where $\langle \cdot | \cdot \rangle$ is the bra-ket notation for inner products and in the last line we have used the definition of an exponential family distribution.
Thus the SP-RM computes the {\em marginal} of the posterior distribution over the target variable, given the input image. This is the conventional approach used in most probabilistic modeling with latent variables.

An alternative and less conventional approach is to use the {\em Max-Sum RM Classifier} (MS-RMC), which \emph{maximizes} over all $g \in \G$ and then chooses the most likely class:
\begin{align} 
	\label{eqn:msc}
	\hat{c}_{\rm MS}(I) &= \argmax_{c \in \Cl}  \max_{g \in \G} p(I|c,g) p(c) p(g) \nonumber\\
	               &= \argmax_{c \in \Cl} \max_{g\in \G} \langle \eta(\theta_{cg}) | T(I) \rangle.
\end{align}

The MS-RMC computes the {\em mode} of the posterior distribution over the target \emph{and} nuisance variables, given the input image. Equivalently, it computes the most likely global configuration of target and nuisance variables for the image. 
Intuitively, this is an effective strategy when there is one explanation $g^* \in \G$ that dominates all other explanations $g \neq g^*$. 
This condition is justified in settings where the rendering function is deterministic or nearly noise-free. 
This approach to classification is unconventional in both the machine learning and computational neuroscience literatures, where the sum-product approach is most commonly used, although it has received some recent attention \cite{bengio2013representation}.
 
Both the sum-product and max-sum classifiers amount to applying an affine transformation to the input image $I$ (via an inner product that performs feature detection via template matching), followed by a sum or max nonlinearity that marginalizes over the nuisance variables. 

Throughout the paper we will assume isotropic or diagonal Gaussian noise for simplicity, but the treatment presented here can be generalized to any distribution from the exponential family in a straightforward manner. Note that such an extension may require a non-linear transformation (i.e. quadratic or logarithmic $T(I)$), depending on the specific exponential family. Please see Supplement Section~\ref{sec:generalize-exp-fam} for more details.

\subsection{Deriving the Key Elements of One Layer of a Deep Convolutional \\ Network from the Rendering Model}
\label{sec:shallowRMC}

Having formulated the Rendering Model (RM), we now show how to connect the RM with deep convolutional networks (DCNs). 
We will see that the MS-RMC (after imposing a few additional assumptions on the RM) gives rise to most commonly used DCN layer types. 

Our first assumption is that the noise added to the rendered template is isotropically Gaussian (GRM) i.e. each pixel has the same noise variance $\sigma^{2}$ that is independent of the configuration $(c,g)$.
Then, assuming the image is normalized $\| I \|^{2}=1$, Eq.~\ref{eqn:msc} yields the {\em max-sum Gaussian RM classifier} (see Appendix~\ref{eq:GRM-MaxOutNN} for a detailed proof)
\begin{align} 
        \label{eqn:cnn1}
\hat{c}_{\rm MS}(I) &= 
\argmax_{c \in \Cl} \max_{g\in \G} \left\langle \frac{1}{\sigma^2} \mu_{cg} \Big| I \right\rangle 
- \frac{1}{2\sigma^2} \|  \mu_{cg}\|_2^2 + \ln\pi_{c}\pi_{g}  \nonumber\\
	&\equiv  \argmax_{c \in \Cl} \max_{g\in \G} \: \langle w_{cg} | I\rangle + b_{cg} ,
\end{align}
where we have defined the \emph{natural parameters} $\eta \equiv \{ w_{cg}, b_{cg} \}$ in terms of the \emph{traditional parameters} $\theta \equiv \{\sigma^2,\mu_{cg}, \pi_c, \pi_g \}$ according to\footnote{Since the Gaussian distribution of the noise is in the exponential family, it can be reparametrized in terms of the natural parameters. This is known as {\em canonical form}.}
\begin{align} 
   w_{cg} &\equiv \frac{1}{\sigma^2} \mu_{cg} = \frac{1}{\sigma^2} R(c,g) \nonumber\\
   b_{cg} &\equiv \frac{1}{2\sigma^2} \|  \mu_{cg}\|_2^2 + \ln\pi_{c}\pi_{g}.
   \label{eqn:grm-weights-biases}
\end{align}
Note that we have suppressed the parameters' dependence on pixel location $x$.

\begin{figure}
   \centering
   \includegraphics[width=\linewidth]{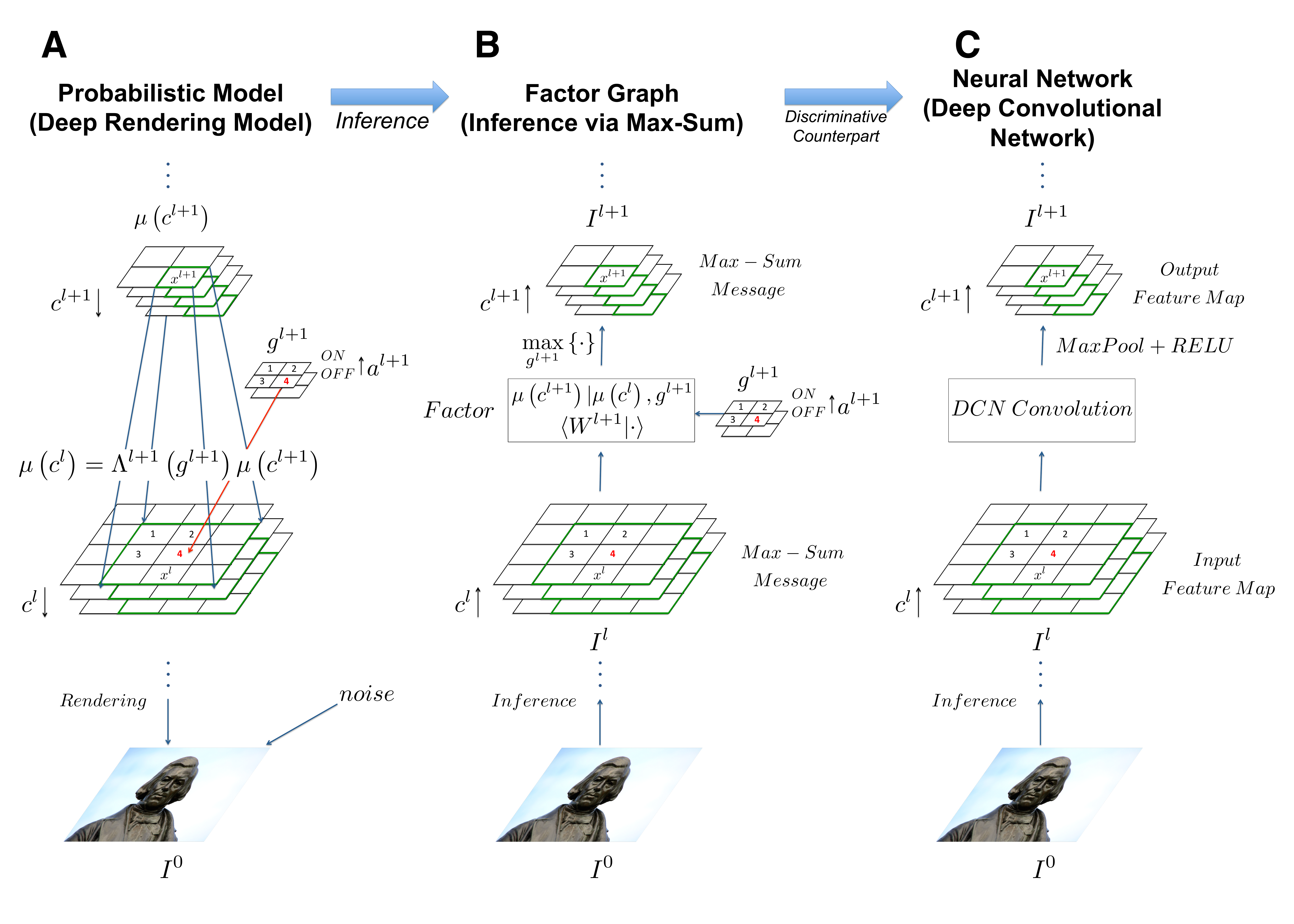} 
   \caption{An example of mapping from the Deep Rendering Model (DRM) to its corresponding factor graph to a Deep Convolutional Network (DCN) showing only the transformation from level $\ell$ of the hierarchy of abstraction to level $\ell+1$.   
   (A) DRM generative model: a single super pixel $x^{\ell+1}$ at level $\ell + 1$ (green, upper) renders down to a $3\times 3$ image patch at level $\ell$ (green, lower), whose location is specified by $g^{\ell+1}$ (red). (B) Factor graph representation of the DRM model that supports efficient inference algorithms such as the max-sum message passing shown here. (C) Computational network that implements the max-sum message passing algorithm from (B) \emph{explicitly}; its structure exactly matches that of a DCN.}
   \label{fig:DRM-to-DCN}
\end{figure}

We will now demonstrate that the sequence of operations in the MS-RMC in Eq.~\ref{eqn:cnn1} coincides exactly with the operations involved in one layer of a DCN (or, more generally, a \emph{max-out neural network}\cite{goodfellow2013maxout}): image normalization, linear template matching, thresholding, and max pooling. See Fig.~\ref{fig:DRM-to-DCN}C.  We now explore each operation in Eq.~\ref{eqn:cnn1} in detail to make the link precise.

First, the image is {\em normalized}. Until recently, there were several different types of normalization typically employed in DCNs: local response normalization, and local contrast normalization\cite{Krizhevsky:2012wl, jarrett2009best}. However, the most recent highly performing DCNs employ a different form of normalization, known as \emph{batch-normalization}\cite{szegedy2014going}. We will come back to this later when we show how to derive batch normalization from a principled approach. One implication of this is that it is unclear what probabilistic assumption the older forms of normalization arise from, if any.

Second, the image is filtered with a set of noise-scaled rendered templates $w_{cg}$. The size of the templates depends on the size of the objects of class $c$ and the values of the nuisance variables $g$. Large objects will have large templates, corresponding to a {\em fully connected layer} in a DCN \cite{lecun1998gradient}, while small objects will have small templates, corresponding to a {\em locally connected layer} in a DCN \cite{wolfdeepface}. If the distribution of objects depends on the pixel position $x$ (e.g., cars are more likely on the ground while planes are more likely in the air) then, in general, we will need different rendered templates at each $x$. In this case, the locally connected layer is appropriate. 
If, on the other hand, all objects are equally likely to be present at all locations throughout the entire image, then we can assume {\it translational invariance} in the RM. This yields a global set of templates that are used at all pixels $x$, corresponding to a {\em convolutional} layer in a DCN \cite{lecun1998gradient} (see Appendix~\ref{lem:trans-to-dcn-conv} for a detailed proof). 
If the filter sizes are large relative to the scale at which the image variation occurs and the filters are overcomplete, then adjacent filters overlap and waste computation. In these case, it is appropriate to use a \emph{strided} convolution, where the output of the traditional convolution is down-sampled by some factor; this saves some computation without losing information.

Third, the resulting activations (log-probabilities of the hypotheses) are passed through a pooling layer; i.e., if $g$ is a translational nuisance, then taking the maximum over $g$ corresponds to {\it max pooling} in a DCN.

Fourth, recall that a given image pixel $x$ will reside in several overlapping image patches, each rendered by its own parent class $c(x)$ and the nuisance location $g(x)$ (Fig.~\ref{fig:DRM-to-DCN}A). Thus we must consider the possibility of {\em collisions}: i.e. when two different parents $c(x_{1}) \neq c(x_{2})$ might render the same pixel (or patch). To avoid such undesirable collisions, it is natural to force the rendering to be {\em locally sparse}: i.e. we must enforce that only one renderer in a local neighborhood can be ``active''. 

To formalize this, we endow each parent renderer with an ON/OFF state via a switching variable $a \in \A \equiv \{\textrm{ON}, \textrm{OFF} \}$. If $a=\textrm{ON}=1$, then the rendered image patch is left untouched, whereas if $a=\textrm{OFF}=0$, the image patch is masked with zeros after rendering. Thus, the switching variable $a$ models (in)active parent renderers. 

However, these switching variables have strong correlations due to the crowding out effect: if one is ON, then its neighbors must be OFF in order to prevent rendering collisions. Although natural for realistic rendering, this complicates inference. Thus, we employ an approximation by instead assuming that the state of each renderer ON or OFF \textit{completely at random} and thus independent of any other variables, including the measurements (i.e., the image itself). Of course, an approximation to real rendering, but it simplifies inference, and leads directly to rectified linear units, as we show below. Such approximations to true sparsity have been extensively studied, and are known as {\em spike-and-slab sparse coding models \cite{lucke2012closed, goodfellow2012large}}.

Since the switching variables are latent (unobserved), we must max-marginalize over them during classification, as we did with nuisance variables $g$ in the last section (one can think of $a$ as just another nuisance). This leads to (see Appendix~\ref{prop:detailedReLuProof} for a more detailed proof)
\begin{align} 
   \label{eqn:relu}
\hat{c}(I) &= \argmax_{c \in \Cl} \max_{g\in \G} \max_{a\in \A} 
	\left\langle \frac{1}{\sigma^2}  a \mu_{cg} \Big| I \right\rangle  -\frac{1}{2\sigma^2} (\| a \mu_{cg} \|_2^{2} + \| I \|_2^{2})  )  + \ln \pi_{c}\pi_{g}\pi_{a} \nonumber\\
		&\equiv \argmax_{c \in \Cl} \max_{g\in \G} \max_{a\in \A} 
		      a(\langle w_{cg} | I\rangle + b_{cg}) +  b_{cga} \nonumber\\
		 &= \argmax_{c \in \Cl} \max_{g\in \G} 
		 \textrm{ReLU}\left(\langle w_{cg} | I\rangle + b_{cg}\right),
\end{align}
where $b_{cga}$ and $b_{cg}$ are bias terms and $\textrm{ReLu}(u) \equiv (u)_+ = \max\{u,0\}$ denotes the soft-thresholding operation performed by the {\it Rectified Linear Units} (ReLU) in modern DCNs \cite{dahl2013improving}. In the last line, we have assumed that the prior $\pi_{cg}$ is uniform so that $b_{cga}$ is independent of $c$ and $g$ and can be dropped.

\subsection{The Deep Rendering Model: Capturing Levels of Abstraction}
\label{sec:drma}

The world is summarizable at varying levels of abstraction, and Hierarchical Bayesian Models (HBMs) can exploit this fact to accelerate learning. In particular, the power of abstraction allows the higher levels of an HBM to learn concepts and categories far more rapidly than lower levels, due to stronger inductive biases and exposure to more data \cite{tenenbaum2011grow}. This is informally known as the \emph{Blessing of Abstraction} \cite{tenenbaum2011grow}. In light of these benefits, it is natural for us to extend the RM into an HBM, thus giving it the power to summarize data at different levels of abstraction.

In order to illustrate this concept, consider the example of rendering an image of a face, at different levels of detail $\ell \in \{L,L-1,\ldots ,0 \}$. At level $\ell=L$ (the coarsest level of abstraction), we specify only the identity of the face $c^L$ and its overall location and pose $g^L$ without specifying any finer-scale details such as the locations of the eyes or type of facial expression. At level $\ell=L-1$, we specify finer-grained details, such as the existence of a left eye ($c^{L-1}$) with a certain location, pose, and state (e.g., $g^{L-1}=$ open or closed),  again without specifying any finer-scale parameters (such as eyelash length or pupil shape). 
We continue in this way, at each level $\ell$ adding finer-scaled information that was unspecified at level $\ell-1$, until at level $\ell=0$ we have fully specified the image's pixel intensities, leaving us with the fully rendered, multi-channel image $I^{0}(x^{\ell},\omega^{\ell})$. Here $x^{\ell}$ refers to a pixel location at level $\ell$.

\begin{figure}
   \centering
   \includegraphics[width=\linewidth]{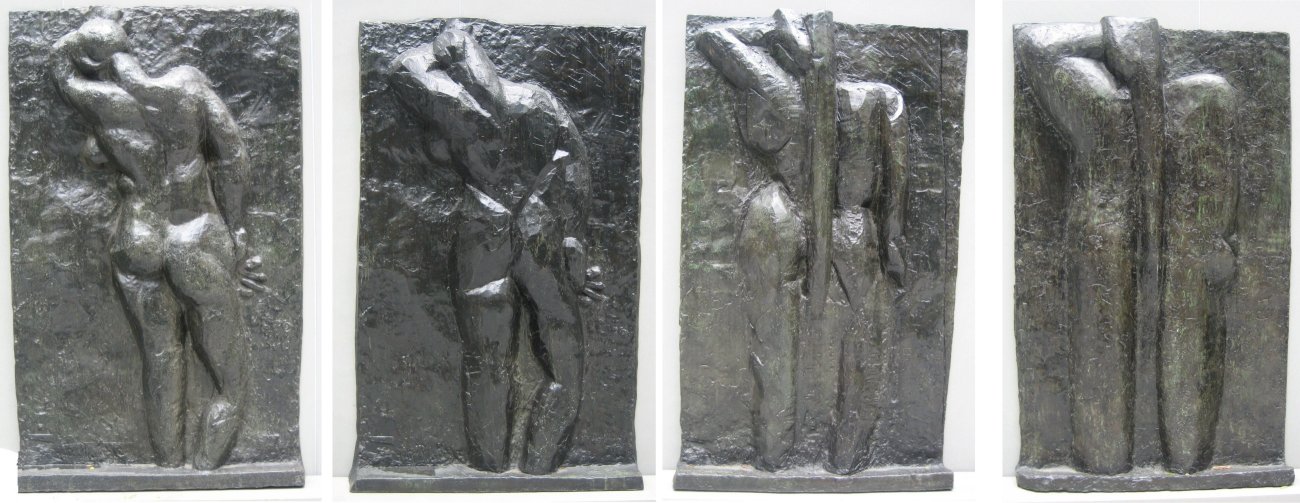}
   \caption{This sculpture by Henri Matisse illustrates the Deep Rendering Model (DRM). The sculpture in the leftmost panel is analogous to a fully rendered image at the lowest abstraction level $\ell=0$. Moving from left to right, the sculptures become progressively more abstract, until the in the rightmost panel we reach the highest abstraction level $\ell=3$. The finer-scale details in the first three panels that are lost in the fourth are the nuisance parameters $g$, whereas the coarser-scale details in the last panel that are preserved are the target $c$.}
   \label{fig:matisse}
\end{figure}

For another illustrative example, consider {\it The Back Series} of sculptures by the artist Henri Matisse (Fig.\ \ref{fig:matisse}). As one moves from left to right, the sculptures become increasingly abstract, losing low-level features and details, while preserving high-level features essential for the overall meaning: i.e. $(c^L,g^L) =$ ``woman with her back facing us.'' Conversely, as one moves from right to left, the sculptures become increasingly concrete, progressively gaining finer-scale details (nuisance parameters $g^\ell, \ell = L-1,\dots,0$) and culminating in a rich and textured rendering.

We formalize this process of progressive rendering by defining the {\em Deep Rendering Model} (DRM). Analogous to the Matisse sculptures, the image generation process in a DRM starts at the highest level of abstraction ($\ell=L$), with the random choice of the object class $c^L$ and overall pose $g^L$. It is then followed by generation of the lower-level details $g^\ell$, and a progressive level-by-level ($\ell \rightarrow \ell-1$) rendering of a set of intermediate rendered ``images'' $\mu^{\ell}$, each with more detailed information. The process finally culminates in a fully rendered $D^{0}\equiv D$-dimensional image $\mu^{0} = I^{0} \equiv I$ ($\ell=0$). Mathematically,
\begin{align} 
	c^{L} &\sim{\rm Cat}(\pi(c^{L})), \quad c^{L} \in \Cl^{L},
	\nonumber \\
	g^{\ell+1} &\sim{\rm Cat}(\pi(g^{\ell+1})), \quad g^{\ell+1} \in \G^{\ell+1}, \quad \ell = L-1, L-2, \dots, 0 	
	\nonumber \\		
	\mu(c^{L}, g) &=  \Lambda(g) \mu(c^{L}) \equiv \Lambda^{1}(g^{1}) \cdots \Lambda^{L}(g^{L}) \cdot \mu(c^{L}), \quad g=\{g^{\ell}\}_{\ell=1}^{L}
	\nonumber \\
	I(c^{L},g) &= \mu(c^{L}, g) + \mathcal{N}(0, \sigma^{2} 1_{D}) \in \R^{D}.
	\label{eqn:DRM}
\end{align}

Here $\Cl^{\ell}, \G^{\ell}$ are the sets of all target-relevant and target-irrelevant nuisance variables at level $\ell$, respectively. The \emph{rendering path} is defined as the sequence $(c^{L},g^{L},\ldots,g^{\ell},\ldots,g^{1})$ from the root (overall class) down to the individual pixels at $\ell=0$.  $\mu(c^{L})$ is an abstract template for the high-level class $c^{L}$, and $\Lambda(g) \equiv \prod_{\ell} \Lambda^{\ell}(g^{\ell})$ represents the sequence of local nuisance transformations that renders finer-scale details as one moves from abstract to concrete. 
Note that each $\Lambda^{\ell}(g^{\ell})$ is an \emph{affine} transformation with a bias term $\alpha(g^{\ell})$ that we have suppressed for clarity\footnote{This assumes that we are using an exponential family with {\em linear sufficient statistics} i.e. $T(I) = (I, 1)^{T}$. However, note that the family we use here is {\em not} Gaussian, it is instead a Factor Analyzer, a different probabilistic model.}.
Figure~\ref{fig:DRM-to-DCN}A illustrates the corresponding graphical model. As before, we have suppressed the dependence of $c^{\ell},g^{\ell}$ on the pixel location $x^\ell$ at level $\ell$ of the hierarchy.

We can cast the DRM into an incremental form by defining an intermediate class $c^{\ell} \equiv (c^{L}, g^{L}, \ldots, g^{\ell+1})$ that intuitively represents a partial rendering path up to level $\ell$. Then, the partial rendering from level $\ell+1$ to $\ell$ can be written as an affine transformation
\begin{align} 
   \mu(c^{\ell}) = \Lambda^{\ell+1}(g^{\ell+1}) \cdot \mu(c^{\ell+1}) + \alpha(g^{\ell+1}) + \mathcal{N}(0, \Psi^{\ell+1}),
   \label{eqn:incDRM}
\end{align}
where we have shown the bias term $\alpha$ explicitly and introduced noise\footnote{We introduce noise for two reasons: (1) it will make it easier to connect later to existing EM algorithms for factor analyzers and (2) we can always take the noise-free limit to impose cluster well-separatedness if needed. Indeed, if the rendering process is deterministic or nearly noise-free, then the latter is justified.} with a diagonal covariance $\Psi^{\ell+1}$. It is important to note that $c^{\ell}, g^\ell$ can correspond to different kinds of target relevant and irrelevant features at different levels.
For example, when rendering faces, $c^1(x^1)$ might correspond to different edge orientations and $g^1(x^1)$ to different edge locations in patch $x^{1}$, whereas $c^2(x^2)$ might correspond to different eye types and $g^2(x^2)$ to different eye gaze directions in patch $x^{2}$.

The DRM generates images at intermediate abstraction levels via the {\em incremental rendering functions} in Eq.~\ref{eqn:incDRM} (see Fig.~\ref{fig:DRM-to-DCN}A). Hence the complete rendering function $R(c,g)$ from Eq.~\ref{eq:render} is a composition of incremental rendering functions, amounting to a product of affine transformations as in Eq.~\ref{eqn:DRM}. Compared to the shallow RM, the factorized structure of the DRM results in an {\it exponential reduction in the number of free parameters}, from $D^{0} \, |\Cl^{L} |\prod_{\ell}|\G_{\ell}|$ to $|\Cl^{L}|\sum_{\ell} D^{\ell} |\G^{\ell}|$ where $D^{\ell}$ is the number of pixels in the intermediate image $\mu^{\ell}$, thus enabling more efficient inference and learning, and most importantly, better generalization.

The DRM as formulated here is distinct from but related to several other hierarchical models, such as the Deep Mixture of Factor Analyzers (DMFA) \cite{tang2012deep} and the Deep Gaussian Mixture Model \cite{van2014factoring}, both of which are essentially compositions of another model --- the Mixture of Factor Analyzers (MFA) \cite{ghahramani1996algorithm}. We will highlight the similarities and differences with these models in more detail in Section~\ref{sec:RelatedWork}.

\subsection{Inference in the Deep Rendering Model} \label{sec:infDRM}

Inference in the DRM is similar to inference in the shallow RM. 
For example, to classify images we can use either the sum-product (Eq.~\ref{eqn:spc}) or the max-sum (Eq.~\ref{eqn:msc}) classifier. 
The key difference between the deep and shallow RMs is that the DRM yields iterated layer-by-layer updates, from fine-to-coarse abstraction (bottom-up) and from coarse-to-fine abstraction (top-down). 
In the case we are only interested in inferring the high-level class $c^{L}$, we only need the fine-to-coarse pass and so we will only consider it in this section. 

Importantly, the bottom-up pass leads directly to DCNs, implying that DCNs ignore potentially useful top-down information. This maybe an explanation for their difficulties in vision tasks with occlusion and clutter, where such top-down information is essential for disambiguating local bottom-up hypotheses. Later on in Section~\ref{sec:top-down-dcn}, we will describe the coarse-to-fine pass and a new class of {\em Top-Down DCNs} that do make use of such information.

Given an input image $I^0$, the max-sum classifier infers the most likely global configuration $\{ c^{\ell}$, $g^\ell \}$, $\ell=0,1,\dots,L$ by executing the max-sum message passing algorithm in two stages: (i) from fine-to-coarse levels of abstraction to infer the overall class label $\hat{c}^L_{\rm MS}$ and (ii) from coarse-to-fine levels of abstraction to infer the latent variables $\hat{c}^\ell_{\rm MS}$ and $\hat{g}^\ell_{\rm MS} $ at all intermediate levels $\ell$. 
As mentioned above, we will focus on the fine-to-coarse pass.
Since the DRM is an RM with a hierarchical prior on the rendered templates, we can use Eq.~\ref{eqn:cnn1} to derive the \emph{fine-to-coarse max-sum DRM classifier (MS-DRMC)} as:
\begin{align} 
	\hat{c}_{MS}(I) &= \argmax_{c^{L} \in \Cl} \max_{g\in \G} \: \langle \eta(c^{L},g) | \Sigma^{-1} | I^{0} \rangle \nonumber\\
	                        &= \argmax_{c^{L} \in \Cl} \max_{g\in \G} \: \langle \Lambda(g) \mu(c^{L}) | (\Lambda(g) \Lambda(g)^{T})^{\dg} | I^{0}  \rangle \nonumber\\
	                        &= \argmax_{c^{L} \in \Cl} \max_{g\in \G} \: \langle \mu(c^{L}) | \prod_{\ell=L}^{1} \Lambda^{\ell}(g^{\ell})^{\dg} | I^{0}  \rangle \nonumber\\
	                        &= \argmax_{c^{L} \in \Cl} \: \langle \mu(c^{L}) | \prod_{\ell=L}^{1} \max_{g^{\ell}\in \G^{\ell}} \Lambda^{\ell}(g^{\ell})^{\dg} | I^{0}  \rangle \nonumber\\
	                        &= \argmax_{c^{L} \in \Cl} \: \langle \mu(c^{L}) | \max_{g^{L}\in \G^{L}} \Lambda^{L}(g^{L})^{\dg} \cdots 
	                            \underbrace{ \max_{g^{1}\in \G^{1}} \Lambda^{1}(g^{1})^{\dg} | I^{0}  }_{\equiv \, I^{1}} \rangle 
	                            \nonumber\\
	                        &\equiv \argmax_{c^{L} \in \Cl} \: \langle \mu(c^{L}) | \max_{g^{L}\in \G^{L}} \Lambda^{L}(g^{L})^{\dg} \cdots 
	                            \underbrace{ \max_{g^{2}\in \G^{2}} \Lambda^{2}(g^{2})^{\dg} | I^{1}  }_{\equiv \, I^{2}}
\rangle 	                            \nonumber\\
	                        &\equiv \argmax_{c^{L} \in \Cl} \: \langle \mu(c^{L}) | \max_{g^{L}\in \G^{L}} \Lambda^{L}(g^{L})^{\dg} \cdots \max_{g^{3}\in \G^{3}} \Lambda^{3}(g^{3})^{\dg} | I^{2}  \rangle \nonumber\\
	                        &~~\vdots \nonumber\\
	                        &\equiv \argmax_{c^{L} \in \Cl} \: \langle \mu(c^{L}) | I^{L}  \rangle, 
	                        \label{eqn:f2c}
\end{align}
where  $\Sigma \equiv \Lambda(g) \Lambda(g)^{T}$ is the covariance of the rendered image $I$ and $\langle x | M | y \rangle \equiv x^{T}M y$. Note the significant change with respect to the shallow RM: the covariance $\Sigma$ is no longer diagonal due to the iterative affine transformations during rendering (Eq.~\ref{eqn:incDRM}), and so we must decorrelate the input  image (via $\Sigma^{-1} I^{0}$  in the first line) in order to classify accurately. 

Note also that we have omitted the bias terms for clarity and that $M^{\dg}$ is the pseudoinverse of matrix $M$. In the fourth line, we used the distributivity of max over products\footnote{ For $a > 0$, $\max \{ab,ac\} = a \max \{b,c\}$.} and in the last lines defined the intermediate quantities
\begin{align} 
	I^{\ell+1} &\equiv \max_{g^{\ell+1}\in \G^{\ell+1}} \langle \underbrace{(\Lambda^{\ell+1}(g^{\ell+1}))^{\dg}}_{\equiv W^{\ell+1}} | I^{\ell} \rangle \nonumber\\
		       &= \max_{g^{\ell+1}\in \G^{\ell+1}} \langle W^{\ell+1}(g^{\ell+1}) | I^{\ell} \rangle \nonumber\\
	               &\equiv {\rm MaxPool}({\rm Conv}(I^{\ell})).
	\label{eqn:iterCG}
\end{align}
Here $I^{\ell}=I^{\ell}(x^{\ell},c^{\ell})$ is the {\em feature map} output of layer $\ell$ indexed by channels $c^{\ell}$ and $\eta(c^{\ell},g^{\ell}) \propto \mu(c^{\ell}, g^{\ell})$ are the natural parameters (i.e., intermediate rendered templates) for level $\ell$. 

If we care only about inferring the overall class of the image $c^L(I^0)$, then the fine-to-coarse pass suffices, since all information relevant to determining the overall class has been integrated. That is, for high-level classification, we need only iterate Eqs.~\ref{eqn:f2c} and \ref{eqn:iterCG}. Note that Eq.~\ref{eqn:f2c} simplifies to Eq.~\ref{eqn:relu} when we assume sparse patch rendering as in Section~\ref{sec:shallowRMC}. 

Coming back to DCNs, we have see that the $\ell$-th iteration of Eq.~\ref{eqn:f2c} or Eq.~\ref{eqn:relu} corresponds to feedforward propagation in the $\ell$-th layer of a DCN. \textit{Thus a DCN's operation has a probabilistic interpretation as fine-to-coarse inference of the most probable global configuration in the DRM.}

\subsubsection{What About the SoftMax Regression Layer?} \label{sec:SoftMaxReg}

It is important to note that we have not fully reconstituted the architecture of modern a DCN as yet.
In particular, the SoftMax regression layer, typically attached to the end of network, is missing. This means that the high-level class $c^{L}$ in the DRM (Eq.~\ref{eqn:f2c}) \emph{is not necessarily the same as the training data class labels} $\tilde{c}$ given in the dataset. In fact, the two labels $\tilde{c}$ and $c^L$ are in general distinct. 

\emph{But then how are we to interpret $c^L$?} The answer is that the most probable global configuration $(c^{L},g^{*})$ inferred by a DCN can be interpreted as a \emph{good representation} of the input image, i.e., one that \emph{disentangles the many nuisance factors into (nearly) independent components} $c^{L},g^{*}$. \footnote{In this sense, the DRM can be seen as a deep (nonlinear) generalization of Independent Components Analysis \cite{hyvarinen2004independent}.} Under this interpretation, it becomes clear that the high-level class $c^L$ in the disentangled representation need not be the same as the training data class label $\tilde{c}$.

The disentangled representation for $c^L$ lies in the penultimate layer activations: $\hat{a}^{L}(I_{n}) = \ln p(c^{L},g^{*}|I_{n})$. Given this representation, we can infer the class label $\tilde{c}$ by using a simple linear classifier such as the SoftMax regression\footnote{Note that this implicitly assumes that a good disentangled representation of an image will be useful for the classification task at hand.}. Explicitly, the Softmax regression layer computes $p(\tilde{c} | \hat{a}^{L}; \theta_{\rm Softmax}) = \phi(W^{L+1} \hat{a}^L + b^{L+1})$, and then chooses the most likely class. Here $\phi(\cdot)$ is the softmax function and $\theta_{\rm Softmax}\equiv\{W^{L+1}, b^{L+1} \}$ are the parameters of the SoftMax regression layer.

\subsection{DCNs are Probabilistic Message Passing Networks}

\subsubsection{Deep Rendering Model and Message Passing}

Encouraged by the correspondence identified in Section \ref{sec:infDRM}, we step back for a moment to reinterpret all of the major elements of DCNs in a probabilistic light. Our derivation of the DRM inference algorithm above is mathematically equivalent to performing \emph{max-sum message passing on the factor graph representation of the DRM}, which is shown in Fig.~\ref{fig:DRM-to-DCN}B. The factor graph encodes the same information as the generative model but organizes it in a manner that simplifies the definition and execution of inference algorithms \cite{kschischang2001factor}. Such inference algorithms are called {\em message passing} algorithms, because they work by passing real-valued functions called messages along the edges between nodes. 
In the DRM/DCN, the messages sent from finer to coarser levels are the feature maps $I^\ell(x^\ell, c^\ell)$. However, unlike the input image $I^{0}$, the channels $c^{\ell}$ in these feature maps do not refer to colors (e.g, red, green, blue) but instead to more abstract features (e.g., edge orientations or the open/closed state of an eyelid).

\subsubsection{A Unification of Neural Networks and Probabilistic Inference}

The factor graph formulation provides a powerful interpretation that the \emph{convolution, Max-Pooling and ReLu operations in a DCN correspond to max-sum inference in a DRM}. 
Thus, we see that architectures and layer types commonly used in today's DCNs are not ad hoc; rather they can be derived from precise probabilistic assumptions that entirely determine their structure. 
Thus the DRM unifies two perspectives --- neural network and probabilistic inference.
A summary of the relationship between the two perspectives is given in Table~\ref{fig:TwoPoVs}.

\begin{table}
   \centering
   \includegraphics[width=0.90\linewidth]{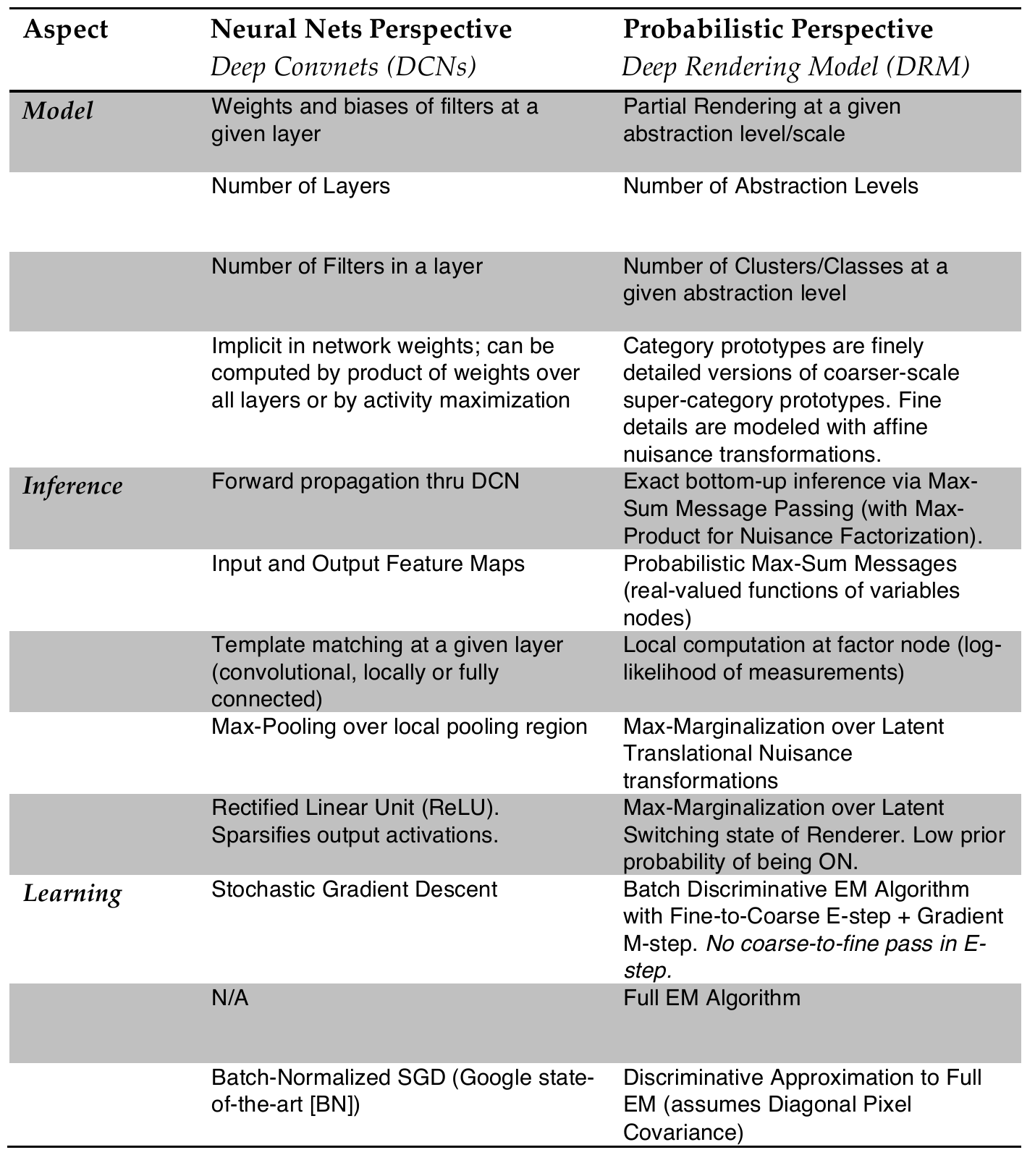} 
   \caption{Summary of probabilistic and neural network perspectives for DCNs. The DRM provides an exact correspondence between the two, providing a probabilistic interpretation for all of the common elements of DCNs relating to the underlying model, inference algorithm, and learning rules. [BN] = reference \cite{ioffe2015batch}.}
   \label{fig:TwoPoVs}
\end{table}

\subsubsection{The Probabilistic Role of Max-Pooling} 

Consider the role of max-pooling from the message passing perspective.  We see that it can be interpreted as the ``max'' in max-sum, thus executing a max-marginalization over nuisance variables $g$. Typically, this operation would be intractable, since there are exponentially many configurations $g \in \G$. But here the DRM's model of abstraction --- a deep product of affine transformations --- comes to the rescue. It enables us to convert an otherwise intractable max-marginalization over $g$ into a tractable sequence of iterated max-marginalizations over abstraction levels $g^{\ell}$ (Eqs.~\ref{eqn:f2c}, \ref{eqn:iterCG}).\footnote{This can be seen, equivalently, as the execution of the max-product algorithm \cite{felzenszwalb2006efficient}.} 
Thus, \emph{the max-pooling operation implements probabilistic marginalization, so is absolutely essential to the DCN's ability to factor out nuisance variation.} 
Indeed, since the ReLu can also be cast as a max-pooling over ON/OFF switching variables, we conclude that \emph{the most important operation in DCNs is max-pooling}. 
This is in conflict with some recent claims to the contrary \cite{hintonVideo}.  

\subsection{Learning the Rendering Models}

Since the RM and DRM are graphical models with latent variables, we can learn their parameters from training data using the expectation-maximization (EM) algorithm \cite{roweis2001learning}. 
We first develop the EM algorithm for the shallow RM from Section \ref{sec:rm} and then extend it to the DRM from Section \ref{sec:drma}.

\subsubsection{EM Algorithm for the Shallow Rendering Model}

Given a dataset of labeled training images $\{ I_n, c_n \}_{n=1}^N$, each iteration of the EM algorithm consists of an E-step that infers the latent variables given the observed variables and the ``old'' parameters $\hat{\theta}_{\rm gen}^{\rm old}$ from the last M-step, followed by an M-step that updates the parameter estimates according to
\begin{align}
\textrm{E-step:}& \quad 
\gamma_{ncg} = p(c, g | I_n; \hat{\theta}_{\rm gen}^{\rm old}), 
\label{eqn:ShallowEStep}
\\[3mm]
\textrm{M-step:}& \quad 
\hat{\theta} = \argmax_{\theta} \sum_n \sum_{cg} \gamma_{ncg} L(\theta).
\label{eqn:EMGeneral}
\end{align}
Here $\gamma_{ncg}$ are the posterior probabilities over the latent mixture components (also called \ the \emph{responsibilities}), the sum $\sum_{cg}$ is over all possible global configurations $(c, g) \in \Cl \times \G$, and $L(\theta)$ is the complete-data log-likelihood for the model.

For the RM, the parameters are defined as $\theta \equiv \{ \pi_{c}, \pi_{g}, \mu_{cg}, \sigma^{2}  \}$ and include the prior probabilities of the different classes $\pi_c$ and nuisance variables $\pi_g$ along with the rendered templates $\mu_{cg}$ and the pixel noise variance $\sigma^2$. If, instead of an isotropic Gaussian RM, we use a full-covariance Gaussian RM or an RM with a different exponential family distribution, then the sufficient statistics and the rendered template parameters would be different (e.g. quadratic for a full covariance Gaussian).

When the clusters in the RM are well-separated (or equivalently, when the rendering introduces little noise), each input image can be assigned to its nearest cluster in a ``hard'' E-step, wherein we care only about the most likely configuration of the $c^\ell$ and $g^\ell$ given the input $I^0$. In this case, the responsibility $\gamma_{ncg}^\ell = 1$ if $c^\ell$ and $g^\ell$ in image $I_{n}$ are consistent with the most likely configuration; otherwise it equals 0. Thus, we can compute the responsibilities using max-sum message passing according to Eqs.~\ref{eqn:f2c} and \ref{eqn:ShallowEStep}.  
In this case, the hard EM algorithm reduces to 
\begin{align} 
   \textrm{ Hard E-step:} \quad \gamma_{ncg}^\ell &= \llbracket (c,g) = (c^*_n,g^*_n) \rrbracket \\[3mm]
    \textrm{M-step:} \quad \hat{N}_{cg}^\ell &= \sum_n \gamma_{ncg}^\ell \nonumber\\
    \hat{\pi}_{cg}^\ell &= \frac{\hat{N}_{cg}^\ell }{N} \nonumber\\
    \hat{\mu}_{cg}^\ell &= \frac{1}{\hat{N}_{cg}^\ell} \sum_n \gamma_{ncg}^\ell I_n^\ell \nonumber\\
    (\hat{\sigma}_{cg}^2)^\ell &= \frac{1}{\hat{N}_{cg}^\ell} \sum_n \gamma_{ncg}^\ell \| I_n^\ell - \mu_{cg}^\ell \|_2^2,
\end{align}
where we have used the Iversen bracket to denote a boolean expression, i.e., $\llbracket b \rrbracket \equiv 1$ if $b$ is true and $\llbracket b \rrbracket \equiv 0$ if $b$ is false.

\subsubsection{EM Algorithm for the Deep Rendering Model}
\label{sec:EM-DRM}

For high-nuisance tasks, the EM algorithm for the shallow RM is computationally intractable, since it requires recomputing the responsibilities and parameters \emph{for all possible configurations} $\tau \equiv ( c^{L}, g^{L},\ldots g^{1})$. 

There are exponentially many such configurations ( $|\Cl^L |\prod_{\ell} |\G^{\ell}|$), one for each possible rendering tree rooted at $c^{L}$. However, the crux of the DRM is the factorized form of the rendered templates (Eq.~\ref{eqn:incDRM}), which results in a dramatic reduction in the number of parameters. This enables us to efficiently infer the most probable configuration \emph{exactly}\footnote{Note that this is exact for the {\em spike-n-slab approximation} to the truly sparse rendering model where only one renderer per neighborhood is active, as described in Section~\ref{sec:shallowRMC}. Technically, this approximation is not a tree, but instead a so-called polytree. Nevertheless, max-sum is exact for trees and polytrees\cite{vamos1992judea}. } via Eq.~\ref{eqn:f2c} and thus avoid the need to resort to slower, approximate sampling techniques (e.g. Gibbs sampling), which are commonly used for approximate inference in deep HBMs \cite{tang2012deep, van2014factoring}. We will exploit this realization below in the DRM E-step.

Guided by the EM algorithm for MFA\cite{ghahramani1996algorithm}, we can extend the EM algorithm for the shallow RM from the previous section into one for the DRM. The DRM E-step performs inference, finding the most likely rendering tree configuration $\tau_n^* \equiv (c_n^L, g_n^L, \ldots, g_n^1)^*$ given the current training input $I_n^0$. The DRM M-step updates the parameters in each layer --- the weights and biases --- via a responsibility-weighted regression of output activations off of input activations. This can be interpreted as each layer learning how to summarize its input feature map into a coarser-grained output feature map, the essence of abstraction. 

In the following it will be convenient to define and use the {\em augmented form}\footnote{$y = mx + b \equiv \tilde{m}^{T} \tilde{x}$, where $\tilde{m} \equiv [ m | b ]$ and $\tilde{x} \equiv [x | 1]$ are the augmented forms for the parameters and input.} for certain parameters so that affine transformations can be recast as linear ones. Mathematically, a single EM iteration for the DRM is then defined as
\begin{align} 
   \textrm{E-step:} \nonumber \\
   	\gamma_{n\tau} &= \llbracket \tau = \tau_{n}^{*} \rrbracket \:\: \textrm{where} \:\: \tau_{n}^{*} \equiv \argmax_{\tau}\left \{\ln p(\tau | I_{n})\right \} \label{eqn:FwdProp} \\[3mm]
   	\Expect \left[ \mu^{\ell}(c^{\ell}) \right] 
	    	&=  \Lambda^{\ell}(g^{\ell})^{\dg}(I^{\ell-1}_{n} - \alpha^{\ell}(g^{\ell})) \equiv W^{\ell}(g^{\ell})I^{\ell-1}_{n} + b^{\ell}(g^{\ell})   \\[3mm]
    	\Expect \left[\mu^{\ell}(c^{\ell})\mu^{\ell}(c^{\ell})^{T} \right] &= \mathbf{1} - \Lambda^{\ell}(g^{\ell})^{\dg}\Lambda^{\ell}(g^{\ell}) + \nonumber\\
	&\qquad \Lambda^{\ell}(g^{\ell})^{\dg}(I^{\ell-1}_{n} - 		\alpha^{\ell}(g^{\ell}))(I^{\ell-1}_{n} - \alpha^{\ell}(g^{\ell}))^{T}(\Lambda^{\ell}(g^{\ell})^{\dg})^{T}  \\[3mm]
    \textrm{M-step:} \nonumber\\
    	\pi(\tau) &= \frac{1}{N} \sum_{n}\gamma_{n\tau} 
	\\[3mm]
    	 \tilde{\Lambda}^{\ell}(g^{\ell}) &\equiv \left[\Lambda^{\ell}(g^{\ell}) \,|\, \alpha^{\ell}(g^{\ell}) \right] 
	 \nonumber\\
    	&= \left(\sum_{n}\gamma_{n\tau}I^{\ell-1}_{n} \Expect \left[ \tilde{\mu}^{\ell}(c^{\ell}) \right]^{T}\right)\left(\sum_{n}\gamma_{n	\tau}\Expect \left[\tilde{\mu}^{\ell}(c^{\ell})\tilde{\mu}^{\ell}(c^{\ell})^{T} \right]\right)^{-1} 
	\\[3mm]
	\Psi^{\ell} &= \frac{1}{N}{\rm diag}\left\{\sum_{n}\gamma_{n\tau}\left(I_{n}^{\ell-1}-\tilde{\Lambda}^{\ell}(g^{\ell})\Expect \left[\tilde{\mu}^{\ell}(c^{\ell}) \right] \right)(I_{n}^{\ell-1})^{T}\right\},	
	\label{eqn:MDRM}
\end{align}
where $\Lambda^{\ell}(g^{\ell})^{\dg} \equiv \Lambda^{\ell}(g^{\ell})^{T}(\Psi^{\ell}+\Lambda^{\ell}(g^{\ell})(\Lambda^{\ell}(g^{\ell}))^{T})^{-1}$ and $\Expect \left[ \tilde{\mu}^{\ell}(c^{\ell}) \right] = \left[\Expect \left[ \mu^{\ell}(c^{\ell}) \right] \,|\ 1\right]$. Note that the nuisance variables $g^{\ell}$ comprise both the translational and the switching variables that were introduced earlier for DCNs.

Note that this new EM algorithm is a \emph{derivative-free alternative to the back propagation algorithm} for training DCNs that is fast, easy to implement, and intuitive. 

A powerful learning rule discovered recently and independently by Google \cite{ioffe2015batch} can be seen as an approximation to the above EM algorithm, whereby Eq.~\ref{eqn:FwdProp} is approximated by normalizing the input activations with respect to each training batch and introducing scaling and bias parameters according to
\begin{align}
	\Expect \left[ \mu^{\ell}(c^{\ell}) \right] 
	    	&=  \Lambda^{\ell}(g^{\ell})^{\dg}(I^{\ell-1}_{n} - \alpha^{\ell}(g^{\ell})) \nonumber\\
		&\approx \Gamma \cdot \tilde{I}^{\ell-1}_n + \beta \nonumber\\
		&\equiv \Gamma \cdot \left( \frac{ I^{\ell-1}_n - \bar{I}_{\mathcal{B}} } {\sigma_{\mathcal{B}}} \right) + \beta.
\end{align}
Here $\tilde{I}^{\ell-1}_n$ are the batch-normalized activations, and $\bar{I}_{\mathcal{B}}$ and $\sigma_{\mathcal{B}}$ are the batch mean and standard deviation vector of the input activations, respectively. Note that the division is element-wise, since each activation is normalized independently to avoid a costly full covariance calculation. The diagonal matrix $\Gamma$ and bias vector $\beta$ are parameters that are introduced to compensate for any distortions due to the batch-normalization. 
In light of our EM algorithm derivation for the DRM, it is clear that this scheme is a crude approximation to the true normalization step in Eq.~\ref{eqn:FwdProp}, whose decorrelation scheme uses the nuisance-dependent mean $\alpha(g^{\ell})$ and full covariance $\Lambda^{\ell}(g^{\ell})^{\dg}$.
Nevertheless, the excellent performance of the Google algorithm bodes well for the performance of the exact EM algorithm for the DRM developed above.

\subsubsection{What About DropOut Training?} \label{sec:DropOut}

We did not mention the most common regularization scheme used with DCNs --- DropOut\cite{hinton2012improving}. DropOut training consists of units in the DCN dropping their outputs at random. This can be seen as a kind of noise corruption, and encourages the learning of features that are robust to missing data and prevents feature co-adaptation as well \cite{hinton2012improving, dahl2013improving}. DropOut is not specific to DCNs; it can be used with other architectures as well. For brevity, we refer the reader to the proof of the DropOut algorithm in Appendix~\ref{prop:GRM-DropOut}. There we show that DropOut can be derived from the EM algorithm.

\subsection{From Generative to Discriminative Classifiers} \label{sec:gen-to-discr}

We have constructed a correspondence between the DRM and DCNs, but the mapping defined so far is not exact. In particular, note the constraints on the weights and biases in Eq.~\ref{eqn:grm-weights-biases}. These are reflections of the distributional assumptions underlying the Gaussian DRM. DCNs do not have such constraints --- their weights and biases are free parameters. As a result, when faced with training data that violates the DRM's underlying assumptions (model misspecification), the DCN will have more freedom to compensate. 
In order to complete our mapping and create an \emph{exact} correspondence between the DRM and DCNs, we relax these parameter constraints, allowing the weights and biases to be free and independent parameters. However, this seems an ad hoc approach. \emph{Can we instead theoretically motivate such a relaxation?}

It turns out that the distinction between the DRM and DCN classifiers is fundamental: the former is known as a \emph{generative classifier} while the latter is known as a \emph{discriminative classifier}\cite{bishop2007generative, jordan2002discriminative}. The distinction between generative and discriminative models has to do with the \emph{bias-variance tradeoff}. On the one hand, generative models have strong distributional assumptions, and thus introduce significant model bias in order to lower model variance (i.e., less risk of overfitting). On the other hand, discriminative models relax some of the distributional assumptions in order to lower the model bias and thus ``let the data speak for itself'', but they do so at the cost of higher variance (i.e., more risk of overfitting)\cite{bishop2007generative, jordan2002discriminative}. Practically speaking, if a generative model is misspecified and if enough labeled data is available, then a discriminative model will achieve better performance on a specific task\cite{jordan2002discriminative}.  However, if the generative model really is the \emph{true} data-generating distribution (or there is not much labeled data for the task), then the generative model will be the better choice. 

Having motivated the distinction between the two types of models, in this section we will define a method for transforming one into the other that we call a \emph{discriminative relaxation}. We call the resulting discriminative classifier a \emph{discriminative counterpart} of the generative classifier.\footnote{The discriminative relaxation procedure is a many-to-one mapping: several generative models might have the same discriminative model as their counterpart.} We will then show that applying this procedure to the generative DRM classifier (with constrained weights) yields the discriminative DCN classifier (with free weights). Although we will focus again on the Gaussian DRM, the treatment can be generalized to other exponential family distributions with a few modifications (see Appendix~\ref{thm:DiscrExpFam} for more details).

\subsubsection{Transforming a Generative Classifier into a Discriminative One}

Before we formally define the procedure, some preliminary definitions and remarks will be helpful. A generative classifier models the joint distribution $p(c,I)$ of the input features \emph{and} the class labels. It can then classify inputs by using Bayes Rule to calculate $p(c | I) \propto p(c,I) = p(I|c)p(c)$ and picking the most likely label $c$. Training such a classifier is known as \emph{generative learning}, since one can generate synthetic features $I$ by sampling the joint distribution $p(c,I)$. Therefore, a generative classifier learns an \emph{indirect} map from input features $I$ to labels $c$ by modeling the joint distribution $p(c,I)$ of the labels \emph{and} the features.

In contrast, a discriminative classifier parametrically models $p(c|I) = p(c| I;\theta_{d})$ and then trains on a dataset of input-output pairs $\{ (I_{n},c_{n}) \}_{n=1}^{N}$ in order to estimate the parameter $\theta_{d}$. This is known as \emph{discriminative learning}, since we directly discriminate between different labels $c$ given an input feature $I$. Therefore, a discriminative classifier learns a direct map from input features $I$ to labels $c$ by \emph{directly} modeling the conditional distribution $p(c| I)$ of the labels \emph{given} the features.

Given these definitions, we can now define the \emph{discriminative relaxation} procedure for converting a generative classifier into a discriminative one. Starting with the standard learning objective for a generative classifier, we will employ a series of transformations and relaxations to obtain the learning objective for a discriminative classifier. Mathematically, we have
\begin{align} 
	\max_{\theta} L_{\rm gen}(\theta) 
	&\equiv \max_{\theta} \sum_{n} \ln p(c_{n}, I_{n} | \theta)
	\nonumber\\ 
	&\overset{(a)}{=} \max_{\theta} \sum_{n} \ln p(c_{n} | I_{n}, \theta) + \ln p(I_{n} | \theta)
	\nonumber\\ 
	&\overset{(b)}{=} \max_{\theta, \tilde{\theta}: \theta = \tilde{\theta} } \sum_{n} \ln p(c_{n} | I_{n}, \theta) + \ln p(I_{n} | \tilde{\theta})
	\nonumber\\ 
	&\overset{(c)}{\leq} \max_{\theta} \underbrace{\sum_{n} \ln p(c_{n} | I_{n}, \theta)}_{\equiv L_{\rm cond}(\theta)}
	\nonumber\\ 
	&\overset{(d)}{=} \max_{\eta : \eta = \rho(\theta)} \sum_{n} \ln p(c_{n} | I_{n}, \eta)
	\nonumber\\ 
	&\overset{(e)}{\leq} \max_{\eta} \underbrace{\sum_{n} \ln p(c_{n} | I_{n}, \eta)}_{\equiv L_{\rm dis}(\eta)},
	\label{eqn:like-gen-cond-discr}
\end{align}
where the $L$'s are the \emph{generative}, \emph{conditional} and \emph{discriminative} log-likelihoods, respectively. In line (a), we used the Chain Rule of Probability. In line (b), we introduced an extra set of parameters $\tilde{\theta}$ while also introducing a constraint that enforces equality with the old set of generative parameters $\theta$. In line (c), we relax the equality constraint (first introduced by Bishop, LaSerre and Minka in \cite{bishop2007generative}), allowing the classifier parameters $\theta$ to differ from the image generation parameters $\tilde{\theta}$. In line (d), we pass to the \emph{natural parametrization} of the exponential family distribution $I | c$, where the natural parameters $\eta = \rho(\theta)$ are a fixed function of the conventional parameters $\theta$. This constraint on the natural parameters ensures that optimization of $L_{\rm cond}(\eta)$ yields the same answer as optimization of $L_{\rm cond}(\theta)$. And finally, in line (e) we relax the natural parameter constraint to get the learning objective for a discriminative classifier, where the parameters $\eta$ are now free to be optimized. 

In summary, starting with a generative classifier with learning objective $L_{\rm gen}(\theta)$, we complete steps (a) through (e) to arrive at a discriminative classifier with learning objective $L_{\rm dis}(\eta)$. We refer to this process as a \emph{discriminative relaxation of a generative classifier} and the resulting classifier is a \emph{discriminative counterpart to the generative classifier}. 

\begin{figure}
   \centering
   \includegraphics[width=0.70\linewidth]{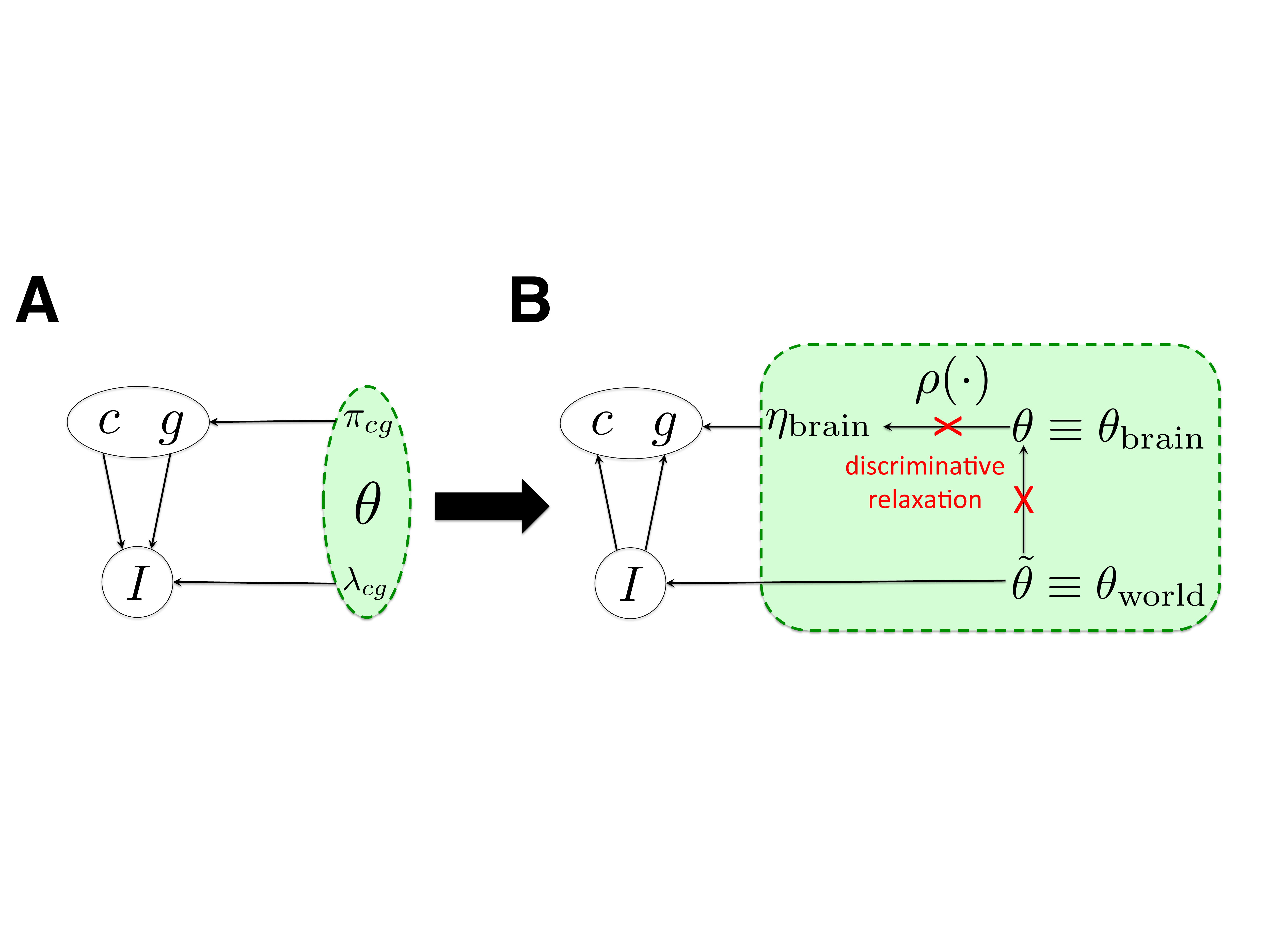} 
   \caption{Graphical depiction of discriminative relaxation procedure. (A) The Rendering Model (RM) is depicted graphically, with mixing probability parameters $\pi_{cg}$ and rendered template parameters $\lambda_{cg}$. The brain-world transformation converts the RM (A) to an equivalent graphical model (B), where an extra set of parameters $\tilde{\theta}$ and constraints (arrows from $\theta$ to $\tilde{\theta}$ to $\eta$) have been introduced. Discriminatively relaxing these constraints (B, red X's) yields the single-layer DCN as the discriminative counterpart to the original generative RM classifier in (A).}
   \label{fig:discr-relax-RM}
\end{figure}

Figure \ref{fig:discr-relax-RM} illustrates the discriminative relaxation procedure as applied to the RM (or DRM). If we consider a Gaussian (D)RM, then $\theta$ simply comprises the mixing probabilities $\pi_{cg}$ and the mixture parameters $\lambda_{cg}$, and so that we have $\theta = \{ \pi_{cg}, \mu_{cg}, \sigma^2 \}$. The corresponding relaxed discriminative parameters are the weights and biases $\eta_{\rm dis} \equiv \{ w_{cg}, b_{cg}\}$.

Intuitively, we can interpret the discriminative relaxation as a {\em brain-world transformation} applied to a generative model. According to this interpretation, instead of the world generating images and class labels (Fig.~\ref{fig:discr-relax-RM}A), we instead imagine the world generating images $I_{n}$ via the rendering parameters $\tilde{\theta} \equiv \theta_{\rm world}$ while the brain generates labels $c_{n},g_{n}$ via the classifier parameters $\eta_{\rm dis} \equiv \eta_{\rm brain}$ (Fig.~\ref{fig:discr-relax-RM}B). Note that the graphical model depicted in Fig.~\ref{fig:discr-relax-RM}B is equivalent to that in Fig.~\ref{fig:discr-relax-RM}A, except for the relaxation of the parameter constraints (red $\times$'s) that represent the discriminative relaxation.

\subsubsection{From the Deep Rendering Model to Deep Convolutional Networks}
\label{sec:dcn-learning-algo}

We can now apply the above to show that \emph{the DCN is a discriminative relaxation of the DRM}. First, we apply the brain-world transformation (Eq.~\ref{eqn:like-gen-cond-discr}) to the DRM. The resulting classifier is precisely a deep \emph{MaxOut neural network \cite{goodfellow2013maxout}} as discussed earlier. Second, we impose translational invariance at the finer scales of abstraction $\ell$ and introduce switching variables $a$ to model inactive renderers. This yields convolutional layers with ReLu activation functions, as in Section~\ref{sec:rm}. Third, the learning algorithm for the generative DRM classifier --- the EM algorithm in Eqs.~\ref{eqn:FwdProp}--\ref{eqn:MDRM} --- must be modified according to Eq.~\ref{eqn:like-gen-cond-discr} to account for the discriminative relaxation. In particular, note that \emph{the new discriminative E-step is only fine-to-coarse and corresponds to forward propagation in DCNs}. As for the discriminative M-step, there are a variety of choices: any general purpose optimization algorithm can be used (e.g., Newton-Raphson, conjugate gradient, etc.). Choosing gradient descent this leads to the classical \emph{back propagation} algorithm for neural network training \cite{wilamowski2001algorithm}. Typically, modern-day DCNs are trained using a variant of back propagation called \emph{Stochastic Gradient Descent (SGD)}, in which gradients are computed using one mini-batch of data at a time (instead of the entire dataset). In light of our developments here, we can re-interpret SGD as a discriminative counterpart to the generative \emph{batch} EM algorithm\cite{cappe2007onlineEM, jordan1998learning}.

This completes the mapping from the DRM to DCNs. We have shown that DCN classifiers are a discriminative relaxation of DRM classifiers, with forward propagation in a DCN corresponding to inference of the most probable configuration in a DRM.\footnote{As mentioned in Section~\ref{sec:SoftMaxReg}, this is typically followed by a Softmax Regression layer at the end. This layer classifies the hidden representation (the penultimate layer activations $\hat{a}^{L}(I_{n})$) into the class labels $\tilde{c}_{n}$ used for training. See Section~\ref{sec:SoftMaxReg} for more details.} 
We have also re-interpreted learning: SGD back propagation training in DCNs is a discriminative relaxation of a batch EM learning algorithm for the DRM. We have provided a principled motivation for passing from the generative DRM to its discriminative counterpart DCN by showing that the discriminative relaxation helps alleviate model misspecification issues by increasing the DRM's flexibility, at the cost of slower learning and requiring more training data.

\section{New Insights into Deep Convolutional Networks}
\label{sec:insights}

In light of the intimate connection between DRMs and DCNs, the DRM provides new insights into how and why DCNs work, answering many open questions. And importantly, DRMs also show us how and why DCNs fail and what we can do to improve them (see Section~\ref{sec:ext}). In this section, we explore some of these insights.

\subsection{DCNs Possess Full Probabilistic Semantics} 

The factor graph formulation of the DRM (Fig.~\ref{fig:DRM-to-DCN}B) provides a useful interpretation of DCNs: it shows us that the convolutional and max-pooling layers correspond to standard message passing operations, as applied \emph{inside} factor nodes in the factor graph of the DRM. 
In particular, \emph{the max-sum algorithm corresponds to a max-pool-conv neural network, whereas the sum-product algorithm corresponds to a mean-pool-conv neural network.} 
More generally, we see that architectures and layer types used commonly in successful DCNs are neither arbitrary nor ad hoc; rather they can be derived from precise probabilistic assumptions that almost entirely determine their structure. 
A summary of the two perspectives --- neural network and probabilistic --- are given in Table \ref{fig:TwoPoVs}.

\subsection{Class Appearance Models and Activity Maximization}

Our derivation of inference in the DRM enables us to understand just how trained DCNs distill and store knowledge from past experiences in their parameters. Specifically, the DRM generates rendered templates $\mu(c^{L}, g) \equiv \mu(c^{L}, g^{L}, \ldots, g^{1})$ via a product of affine transformations, thus implying that \emph{class appearance models in DCNs (and DRMs) are stored in a factorized form across multiple levels of abstraction.} Thus, we can explain why past attempts to understand how DCNs store memories by examining filters at each layer were a fruitless exercise: it is \emph{the product of all the filters/weights over all layers that yield meaningful images of objects}. Indeed, this fact is encapsulated mathematically in Eqs.~\ref{eqn:DRM}, \ref{eqn:incDRM}. Notably, recent studies in computational neuroscience have also shown a strong similarity between representations in primate visual cortex and a highly trained DCN \cite{yamins2014performance}, suggesting that the brain might also employ factorized class appearance models.

\begin{figure}
   \centering
   \includegraphics[height=0.80\textheight]{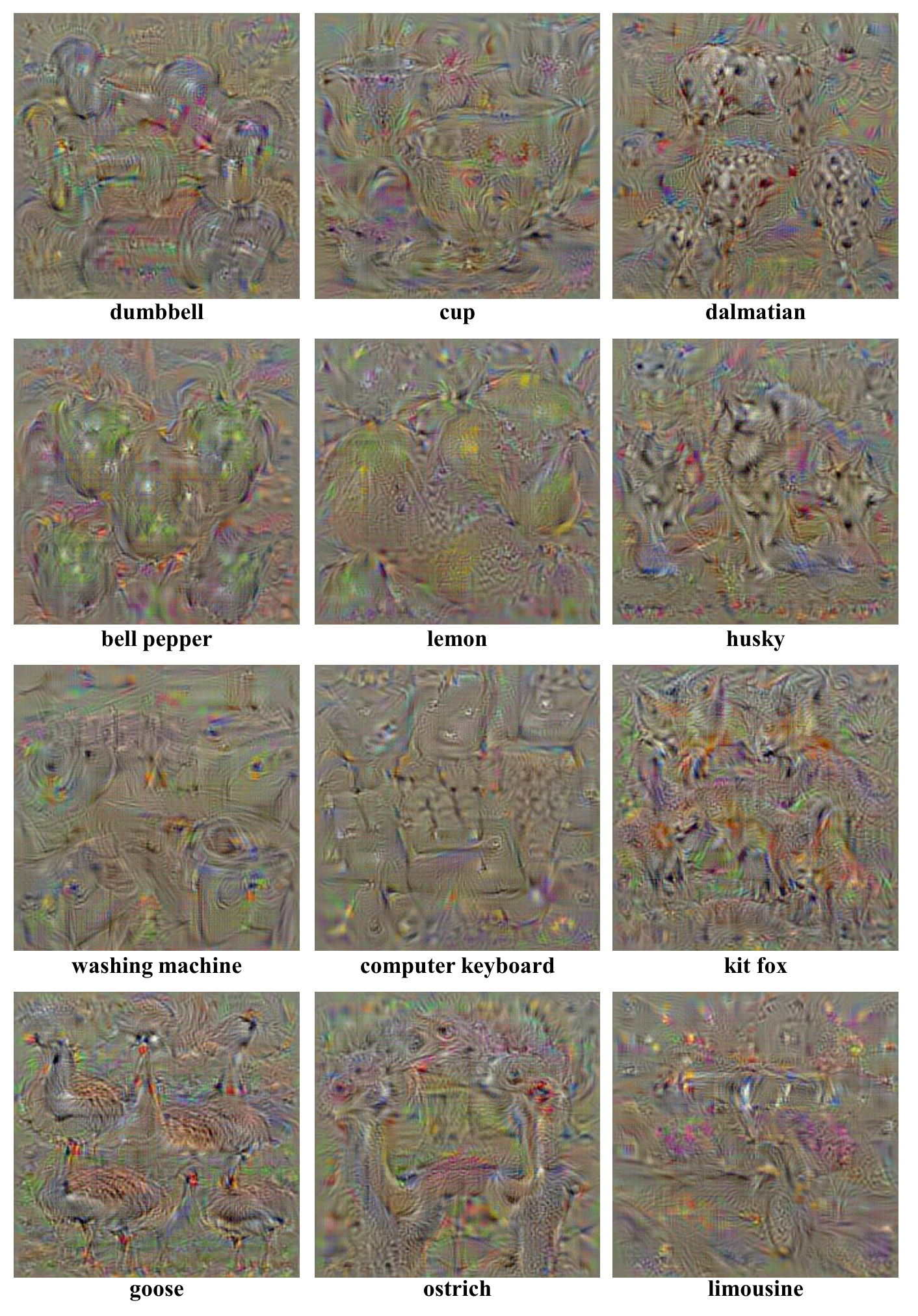} 
   \caption{Results of activity maximization on ImageNet dataset. 
   For a given class $c$, activity-maximizing inputs are superpositions of various poses of the object, with distinct patches $\Pa_{i}$ containing distinct poses $g^{*}_{\Pa_{i}}$, as predicted by Eq. \ref{eqn:actmax}. Figure adapted from \cite{simonyan2013deep} with permission from the authors.}
   \label{fig:actmax}
\end{figure}

We can also shed new light on another approach to understanding DCN memories that proceeds by searching for input images that maximize the activity of a particular class unit (say, cat) \cite{simonyan2013deep}, a technique we call \emph{activity maximization}. Results from activity maximization on a high performance DCN trained on 15 million images from \cite{simonyan2013deep} is shown in Fig.~\ref{fig:actmax}. The resulting images are striking and reveal much about how DCNs store memories.   We now derive a closed-form expression for the activity-maximizing images as a function of the underlying DRM model's learned parameters. Mathematically, we seek the image $I$ that maximizes the score $S(c|I)$ of a specific object class. Using the DRM, we have
\begin{align} 
	 \max_{I} S(c^{\ell} | I) &=\max_{I} \max_{g\in \G} \: \langle \frac{1}{\sigma^{2}}\mu(c^{\ell},g^{\ell}) | I \rangle \nonumber\\
	 		         &\propto \max_{g\in \G} \max_{I} \: \langle \mu(c^{\ell},g) | I \rangle 
			         \nonumber\\
				&= \max_{g\in \G} \max_{I_{\Pa_{1}}} \cdots \max_{I_{\Pa_{p}}} \: \langle \mu(c^{\ell},g) | \sum_{\Pa_{i} \in \Pa} I_{\Pa_{i}} \rangle
				\nonumber\\
				&= \max_{g\in \G} \: \sum_{\Pa_{i} \in \Pa} \max_{I_{\Pa_{i}}} \: \langle \mu(c^{\ell},g) |  I_{\Pa_{i}} \rangle 
				\nonumber\\
				&= \max_{g\in \G} \: \sum_{\Pa_{i} \in \Pa} \: \langle \mu(c^{\ell},g) |  I^{*}_{\Pa_{i}}(c^{\ell},g) \rangle 
				\nonumber\\
				&= \sum_{\Pa_{i} \in \Pa} \: \langle \mu(c^{\ell},g) |  I^{*}_{\Pa_{i}}(c^{\ell},g^{*}_{\Pa_{i}} \rangle ,
\end{align}
\sloppy
where $I^{*}_{\Pa_{i}}(c^{\ell},g) \equiv \argmax_{I_{\Pa_{i}}} \: \langle \mu(c^{\ell},g) | I_{\Pa_{i}} \rangle$ and $g^{*}_{\Pa_{i}}=g^{*}(c^{\ell},\Pa_{i}) \equiv \argmax_{g \in \G} \: \langle \mu(c^{\ell},g) | I^{*}_{\Pa_{i}}(c^{\ell},g)  \rangle$. In the third line, the image $I$ is decomposed into $P$ patches $I_{\Pa_{i}}$ of the same size as $I$, with all pixels outside of the patch $\Pa_{i}$ set to zero. The $\max_{g \in \G}$ operator finds the most probable $g^{*}_{\Pa_{i}}$ within each patch. The solution $I^{*}$ of the activity maximization is then the sum of the individual activity-maximizing patches
\begin{align} 
	I^{*} \equiv \sum_{\Pa_{i} \in \Pa} I^{*}_{\Pa_{i}}(c^{\ell},g^{*}_{\Pa_{i}}) \propto \sum_{\Pa_{i} \in \Pa} \mu(c^{\ell},g^{*}_{\Pa_{i}}).  \label{eqn:actmax}									
\end{align}
Eq.~\ref{eqn:actmax} implies that $I^{*}$ contains multiple appearances of the same object but in various poses. Each activity-maximizing patch has its own pose (i.e. $g^{*}_{\Pa_{i}}$), in agreement with Fig.~\ref{fig:actmax}. Such images provide strong confirming evidence that the underlying model is a mixture over nuisance (pose) parameters, as is expected in light of the DRM.

\subsection{(Dis)Entanglement: Supervised Learning of Task Targets Is \\ Intertwined with Unsupervised Learning of Latent Task Nuisances}

A key goal of representation learning is to disentangle the factors of variation that contribute to an image's appearance. Given our formulation of the DRM, it is clear that DCNs are discriminative classifiers that capture these factors of variation with latent nuisance variables $g$. As such, the theory presented here makes a clear prediction that \emph{for a DCN, supervised learning of task targets will lead inevitably to unsupervised learning of latent task nuisance variables}. From the perspective of manifold learning, this means that the architecture of DCNs is designed to learn and disentangle the intrinsic dimensions of the data manifold.

\begin{figure}
   \centering
   \includegraphics[width=4.1in]{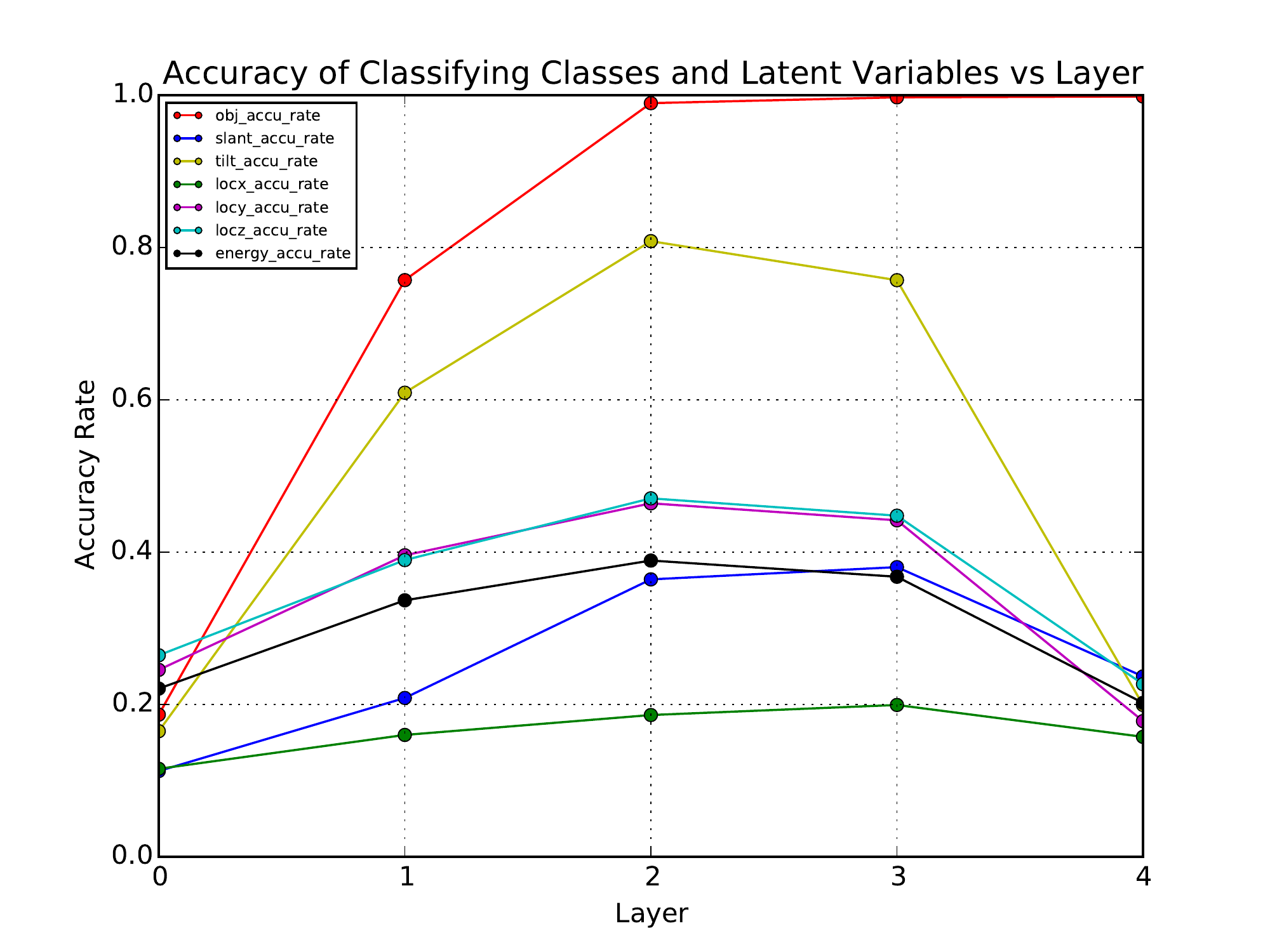} 
   \includegraphics[width=4.1in]{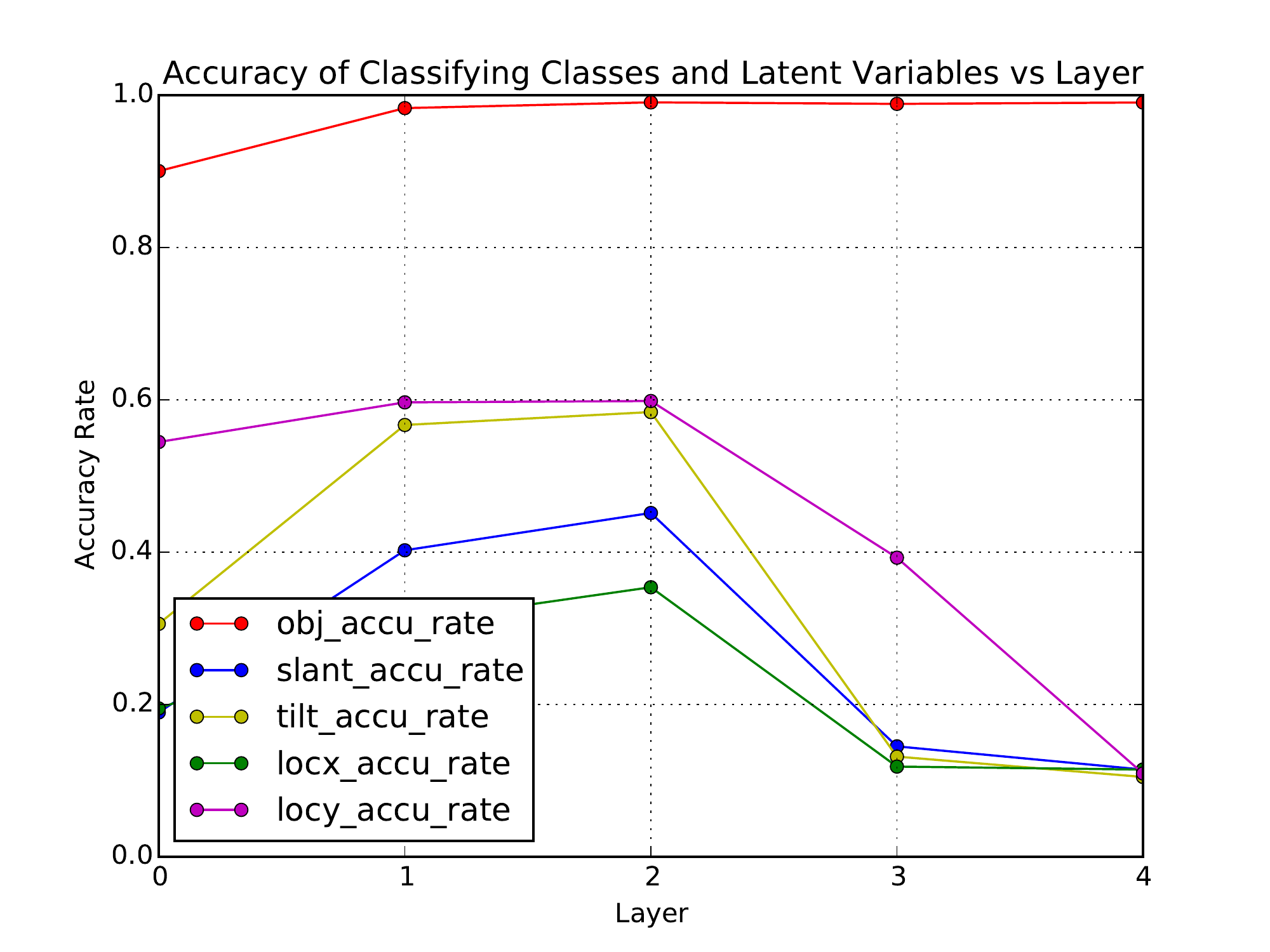} 
   \caption{Manifold entanglement and disentanglement as illustrated in a 5-layer max-out DCN trained to classify synthetically rendered images of planes (top) and naturalistic objects (bottom) in different poses, locations, depths and lighting conditions. The amount of linearly separable information about the target variable (object identity, red) increases with layer depth while information about nuisance variables (slant, tilt, left-right location, depth location) follows an inverted U-shaped curve. Layers with increasing information correspond to disentanglement of the manifold --- factoring variation into independent parameters --- whereas layers with decreasing information correspond to marginalization over the nuisance parameters. Note that disentanglement of the latent nuisance parameters is achieved progressively over multiple layers, without requiring the network to explicitly train for them. Due to the complexity of the variation induced, several layers are required for successful disentanglement, as predicted by our theory. }
   \label{fig:entanglement}
\end{figure}

In order to test this prediction, we trained a DCN to classify synthetically rendered images of naturalistic objects, such as cars and planes. Since we explicitly used a renderer, we have the power to systematically control variation in factors such as pose, location, and lighting. After training, we probed the layers of the trained DCN to quantify how much linearly separable information exists about the task target $c$ and latent nuisance variables $g$.  Figure~\ref{fig:entanglement} shows that the trained DCN possesses significant information about latent factors of variation and, furthermore, the more nuisance variables, the more layers are required to disentangle the factors. This is strong evidence that depth is necessary and that the amount of depth required increases with the complexity of the class models and the nuisance variations.

In light of these results, when we talk about training DCNs, the traditional distinction between supervised and unsupervised learning is ill-defined at worst and misleading at best.  This is evident from the initial formulation of the RM, where $c$ is the task target and $g$ is a latent variable capturing all nuisance parameters (Fig.~\ref{fig:bigFig}). Put another way, our derivation above shows that \emph{DCNs are discriminative classifiers with latent variables that capture nuisance variation}. We believe the main reason this was not noticed earlier is probably that latent nuisance variables in a DCN are hidden within the max-pooling units, which serve the dual purpose of learning \emph{and} marginalizing out the latent nuisance variables.

\section{From the Deep Rendering Model to Random Decision Forests} 
\label{sec:rdf}

Random Decision Forests (RDF)s \cite{criminisi2013decision, breiman2001random} are one of the best performing but least understood classifiers in machine learning. While intuitive, their structure does not seem to arise from a proper probabilistic model. Their success in a vast array of ML tasks is perplexing, with no clear explanation or theoretical understanding. In particular, they have been quite successful in real-time image and video segmentation tasks, the most prominent example being their use for pose estimation and body part tracking in the Microsoft Kinect gaming system \cite{pham2015study}. They also have had great success in medical image segmentation problems \cite{criminisi2013decision, breiman2001random}, wherein distinguishing different organs or cell types is quite difficult and typically requires expert annotators.

In this section we show that, like DCNs, RDFs can also be derived from the DRM model, but with a different set of assumptions regarding the nuisance structure. Instead of translational and switching nuisances, we will show that an {\em additive mutation nuisance process} that generates a hierarchy of categories (e.g., evolution of a taxonomy of living organisms) is at the heart of the RDF. 
As in the DRM to DCN derivation, we will start with a generative classifier and then derive its discriminative relaxation. As such, RDFs possess a similar interpretation as DCNs in that they can be cast as max-sum message passing networks.

A \emph{decision tree classifier} takes an input image $I$ and asks a series of questions about it. The answer to each question determines which branch in the tree to follow. At the next node, another question is asked. This pattern is repeated until a leaf $b$ of the tree is reached. At the leaf, there is a class posterior probability distribution $p(c | I, b)$ that can be used for classification. Different leaves contain different class posteriors. 
An RDF is an ensemble of decision tree classifiers $t \in \mathcal{T}$. 
To classify an input $I$, it is sent as input to each decision tree $t \in \mathcal{T}$ individually, and each decision tree outputs a class posterior $p(c | I, b, t)$. These are then averaged to obtain an overall posterior $p(c | I) = \sum_{t} p(c | I, b, t)p(t)$, from which the most likely class $c$ is chosen. Typically we assume $p(t) = 1/|\mathcal{T}|$.

\subsection{The Evolutionary Deep Rendering Model: A Hierarchy of Categories}

We define the \emph{evolutionary DRM} (E-DRM) as a DRM with an evolutionary tree of categories. Samples from the model are generated by starting from the root ancestor template and randomly mutating the templates. Each child template is an additive mutation of its parent, where the specific mutation does not depend on the parent (see Eq.~\ref{eqn:incDRM-RDF} below). 
Repeating this pattern at each child node, an entire evolutionary tree of templates is generated. We assume for simplicity that we are working with a Gaussian E-DRM so that at the leaves of the tree a sample is generated by adding Gaussian pixel noise. Of course, as described earlier, this can be extended to handle other noise distributions from the exponential family. Mathematically, we have
\begin{align} 
	c^{L} &\sim{\rm Cat}(\pi(c^{L})), \quad c^{L} \in \Cl^{L},	
	\nonumber \\[2mm]
	g^{\ell+1} &\sim{\rm Cat}(\pi(g^{\ell+1})), \quad g^{\ell+1} \in \G^{\ell+1}, \quad \ell = L-1, L-2, \dots, 0 	 
	\nonumber \\[2mm]	
	\mu(c^{L}, g) &=  \Lambda(g) \mu(c^{L}) \equiv \Lambda^{1}(g^{1}) \cdots \Lambda^{L}(g^{L}) \cdot \mu(c^{L})
	\nonumber \\
	&= \mu(c^{L}) + \alpha(g^{L}) + \cdots + \alpha(g^{1}), \quad g=\{g^{\ell}\}_{\ell=1}^{L}
	\nonumber \\[2mm]
	I(c^{L},g) &= \mu(c^{L}, g) + \mathcal{N}(0, \sigma^{2} 1_{D}) \in \R^{D}.
	\label{eqn:DRM-RDF}
\end{align}
Here, $\Lambda^{\ell}(g^{\ell})$ has a special structure due to the additive mutation process: $\Lambda^{\ell}(g^{\ell}) = [ {\bf 1} \, |\, \alpha(g^{\ell}) ]$, where ${\bf 1}$ is the identity matrix. As before, $\Cl^{\ell}, \G^{\ell}$ are the sets of all target-relevant and target-irrelevant nuisance variables at level $\ell$, respectively. (The target here is the same as with the DRM and DCNs --- the overall class label $c^L$.) The rendering path represents template evolution and is defined as the sequence $(c^{L},g^{L},\ldots,g^{\ell},\ldots,g^{1})$ from the root ancestor template down to the individual pixels at $\ell=0$.  $\mu(c^{L})$ is an abstract template for the root ancestor $c^{L}$, and $\sum_{\ell} \alpha(g^{\ell})$ represents the sequence of local nuisance transformations, in this case, the accumulation of many additive mutations. 

As with the DRM, we can cast the E-DRM into an incremental form by defining an intermediate class $c^{\ell} \equiv (c^{L}, g^{L}, \ldots, g^{\ell+1})$ that intuitively represents a partial evolutionary path up to level $\ell$. Then, the mutation from level $\ell+1$ to $\ell$ can be written as
\begin{align} 
   \mu(c^{\ell}) = \Lambda^{\ell+1}(g^{\ell+1}) \cdot \mu(c^{\ell+1}) = \mu(c^{\ell+1}) + \alpha(g^{\ell + 1}),
   \label{eqn:incDRM-RDF}
\end{align}
where $\alpha(g^{\ell})$ is the mutation added to the template at level $\ell$ in the evolutionary tree.

As a generative model, the E-DRM is a {\em mixture of evolutionary paths}, where each path starts at the root and ends at a leaf species in the tree. Each leaf species is associated with a rendered template $\mu(c^{L},g^{L},\ldots, g^{1})$.

\subsection{Inference with the E-DRM Yields a Decision Tree}

Since the E-DRM is an RM with a hierarchical prior on the rendered templates, we can use Eq.~\ref{eqn:cnn1} to derive the E-DRM inference algorithm as:
\begin{align} 
	\hat{c}_{MS}(I) &= \argmax_{c^{L} \in \Cl^{L}} \max_{g\in \G} \: \langle \eta(c^{L},g) | I^{0} \rangle \nonumber\\
	                        &= \argmax_{c^{L} \in \Cl^{L}} \max_{g\in \G} \: \langle \Lambda(g) \mu(c^{L}) | I^{0}  \rangle \nonumber\\
	                        &= \argmax_{c^{L} \in \Cl^{L}} \max_{g^{1}\in \G^{1}} \cdots \max_{g^{L}\in \G^{L}} \: \langle \mu(c^{L}) + \alpha(g^{L}) + \cdots + \alpha(g^{1}) | I^{0}  \rangle \nonumber\\
	                        &= \argmax_{c^{L} \in \Cl^{L}} \max_{g^{1}\in \G^{1}} \cdots \max_{g^{L-1}\in \G^{L-1}} \: 
	                        		\langle \underbrace{\mu(c^{L}) + \alpha(g^{L*})}_{\equiv \mu(c^{L},g^{L*}) = \mu(c^{L-1})} + \cdots + \alpha(g^{1}) | I^{0}  
					\rangle \nonumber\\
	                        &= \argmax_{c^{L} \in \Cl^{L}} \max_{g^{1}\in \G^{1}} \cdots \max_{g^{L-1}\in \G^{L-1}} \: 
	                        		\langle \mu(c^{L-1}) + \alpha(g^{L-1}) + \cdots + \alpha(g^{1}) | I^{0}  
					\rangle \nonumber\\
	                        &~~\vdots \nonumber\\
	                        &\equiv \argmax_{c^{L} \in \Cl^{L}} \: \langle \mu(c^{L}, g^{*}) | I^{0}  \rangle,                 \label{eqn:inf-DRM-RDF}
\end{align}
Note that we have explicitly shown the bias terms here, since they represent the additive mutations. In the last lines, we repeatedly use the distributivity of max over sums, resulting in the iteration
\begin{align}
	g^{\ell+1}(c^{\ell+1})^{*} &\equiv \argmax_{g^{\ell+1}\in \G^{\ell+1}} \langle \underbrace{\mu(c^{\ell+1},g^{\ell+1})}_{\equiv W^{\ell+1}} | I^{0} \rangle \nonumber\\
		       &= \argmax_{g^{\ell+1}\in \G^{\ell+1}} \langle W^{\ell+1}(c^{\ell+1},g^{\ell+1}) | I^{0} \rangle \nonumber\\
	               &\equiv {\rm ChooseChild}({\rm Filter}(I^{0})).
	\label{eqn:iterCG-RDF}
\end{align}

Note the key differences from the DRN/DCN inference derivation in Eq.~\ref{eqn:f2c}: (i) the input to each layer is always the input image $I^{0}$, (ii) the iterations go from coarse-to-fine (from root ancestor to leaf species) rather than fine-to-coarse, and (iii) the resulting network is not a neural network but rather a deep decision tree of single-layer neural networks. These differences are due to the special additive structure of the mutational nuisances and the evolutionary tree process underlying the generation of category templates.

\subsubsection{What About the Leaf Histograms?}

The mapping to a single decision tree is not yet complete; the leaf label histograms \cite{criminisi2013decision, breiman2001random} are missing. Analogous to the missing SoftMax regression layers with DCNs (Sec~\ref{sec:SoftMaxReg}), the high-level representation class label $c^{L}$ inferred by the E-DRM in Eq.~\ref{eqn:inf-DRM-RDF} \emph{need not be the training data class label} $\tilde{c}$. For clarity, we treat the two as separate in general. 

But then how do we understand $c^L$? We can interpret the inferred configuration $\tau^{*} = (c^{L*}, g^{*})$ as \emph{a disentangled representation} of the input, wherein the different factors in $\tau^*$, including $c^L$, vary independently in the world. In contrast to DCNs, the class labels $\tilde{c}$ in a decision tree are instead inferred from the \emph{discrete evolutionary path} variable $\tau^{*}$ through the use of the leaf histograms $p(\tilde{c} | \tau^{*})$. Note that decision trees also have label histograms at all internal (non-leaf) nodes, but that they are not needed for inference. However, they do play a critical role in learning, as we will see below.

We are almost finished with our mapping from inference in Gaussian E-DRMs to decision trees. To finish the mapping, we need only apply the discriminative relaxation (Eq.~\ref{eqn:like-gen-cond-discr}) in order to allow the weights and biases that define the decision functions in the internal nodes to be free. Note that this is exactly analogous to steps in Section~\ref{sec:gen-to-discr} for mapping from the Gaussian DRM to DCNs.

\subsection{Bootstrap Aggregation to Prevent Overfitting Yields A Decision \\ Forest}

Thus far we have derived the inference algorithm for the E-DRM and shown that its discriminative counterpart is indeed a single decision tree. But how to relate to this result to the entire forest? This is important, since it is well known that individual decision trees are notoriously good at overfitting data. Indeed, the historical motivation for introducing a forest of decision trees has been in order to prevent such overfitting by averaging over many different models, each trained on a randomly drawn subset of the data. This technique is known as \emph{bootstrap aggregation} or \emph{bagging} for short, and was first introduced by Breiman in the context of decision trees \cite{breiman2001random}. For completeness, in this section we review bagging, thus completing our mapping from the E-DRM to the RDF.

In order to derive bagging, it will be necessary in the following to make explicit the dependence of learned inference parameters $\theta$ on the training data $\D_{CI} \equiv \{ (c_{n}, I_{n}) \}_{n=1}^{N}$, i.e. $\theta = \theta(\D_{CI})$. This dependence is typically suppressed in most work, but is necessary here as bagging entails \emph{training different decision trees $t$ on different subsets $\D_t \subset \D$ of the full training data}. In other words, $\theta_t = \theta_t(\D_t)$. 

Mathematically, we perform inference as follows: Given all previously seen data $\D_{CI}$ and an unseen image $I$, we classify $I$ by computing the posterior distribution
\begin{align} 
   p(c | I, \D_{CI}) &= \sum_{A} p(c, A | I, \D_{CI})     \nonumber\\
   		           &= \sum_{A} p(c | I, \D_{CI}, A) p(A)     \nonumber\\
   		           &\equiv \Expect_{A} [ p(c | I, \D_{CI}, A) ]     \nonumber\\
   		           &\overset{(a)}{\approx} \frac{1}{T}\sum_{t \in \T} p(c | I, \D_{CI}, A_{t})     \nonumber\\
   		           &\overset{(b)}{=} \frac{1}{T}\sum_{t \in \T} \int d \theta_{t} \, p(c | I,\theta_{t}) \, p(\theta_{t} | \D_{CI}, A_{t})     \nonumber\\
   		           &\overset{(c)}{\approx} 
		                \underbrace{ \frac{1}{T}\sum_{t \in \T} p(c | I,\theta_{t}^{*})}_{\rm Decision\, Forest\, Classifier}, \quad \theta_{t}^{*} \equiv  \max_{\theta} p(\theta | \D_{CI}(A_{t})).
\end{align}
Here $A_{t} \equiv (a_{tn})_{n=1}^{N}$ is a collection of switching variables that indicates which data points are included, i.e., $a_{tn} = 1$ if data point $n$ is included in dataset $\D_t \equiv \D_{CI}(A_{t})$. In this way, we have randomly subsampled the full dataset $\D_{CI}$ (with replacement) $T$ times in line (a), approximating the true marginalization over all possible subsets of the data. In line (b), we perform Bayesian Model Averaging over all possible values of the E-DRM/decision tree parameters $\theta_{t}$. Since this is intractable, we approximate it with the MAP estimate $\theta_{t}^{*}$ in line (c). The overall result is that each E-DRM (or decision tree) $t$ is trained separately on a randomly drawn subset $\D_t \equiv \D_{CI}(A_{t}) \subset \D_{CI}$ of the entire dataset, and the final output of the classifier is an average over the individual classifiers. 

\subsection{EM Learning for the E-DRM Yields the InfoMax Principle}

One approach to train an E-DRM classifier is to maximize the mutual information between the given training labels $\tilde{c}$ and the inferred (partial) rendering path $\tau^{\ell} \equiv (c^{L},g^{L},\ldots,g^{l})$ at each level. Note that $\tilde{c}$ and $\tau^{\ell}$ are both \emph{discrete} random variables.

This Mutual Information-based Classifier (MIC) plays the same role as the Softmax regression layer in DCNs, predicting the class labels $\tilde{c}$ given a good disentangled representation $\tau^{\ell *}$ of the input $I$. In order to train the MIC classifier, we update the classifier parameters $\theta_{\rm MIC}$ in each M-step as the solution to the optimization: 
\begin{align}
   \max_{\theta} MI(\tilde{c},(c^{L},g^{L},\ldots,g^{1}))
   & = \max_{\theta^{1}} \cdots \max_{\theta^{L}} \sum_{l=L}^{1} MI(\tilde{c},g_{n}^{l} | g_{n}^{l+1};\theta^{l})    \nonumber\\
   & = \sum_{l=L}^{1} \max_{\theta^{l}} MI(\tilde{c},g_{n}^{l} | g_{n}^{l+1};\theta^{l}) \nonumber\\
   & = \sum_{l=L}^{1} \max_{\theta^{l}} \underbrace{H[\tilde{c}] - H[\tilde{c} | g_{n}^{l};\theta^{l}]}_{\equiv \textrm{Information Gain}}.  
\end{align}
Here $MI(\cdot, \cdot)$ is the \emph{mutual information} between two random variables, $H[\cdot]$ is the entropy of a random variable, and $\theta^{\ell}$ are the parameters at layer $\ell$. In the first line, we have used the layer-by-layer structure of the E-DRM to split the mutual information calculation across levels, from coarse to fine. In the second line, we have used the max-sum algorithm (dynamic programming) to split up the optimization into a sequence of optimizations from $\ell = L \rightarrow \ell=1$. In the third line, we have used the information-theoretic relationship $MI(X,Y) \equiv H[X] - H[Y|X]$. This algorithm is known as \emph{InfoMax} in the literature \cite{criminisi2013decision}.

\section{Relation to Prior Work} \label{sec:RelatedWork}

\subsection{Relation to Mixture of Factor Analyzers}

As mentioned above, on a high level, the DRM is related to hierarchical models based on the Mixture of Factor Analyzers (MFA) \cite{ghahramani1996algorithm}. Indeed, if we add noise to each partial rendering step from level $\ell$ to $\ell-1$ in the DRM, then Eq.~\ref{eqn:incDRM} becomes
\begin{align}
	I^{\ell-1} \sim  \mathcal{N}\left(\Lambda^{\ell}(g^{\ell}) \mu^{\ell}(c^{\ell}) +  \alpha^{\ell}(g^{\ell}), \Psi^{\ell}\right),
\end{align}
where we have introduced the diagonal noise covariance $\Psi^{\ell}$. This is equivalent to the MFA model. 
The DRM and DMFA both employ parameter sharing, resulting in an exponential reduction in the number of parameters, as compared to the collapsed or shallow version of the models. This serves as a strong regularizer to prevent overfitting.

Despite the high-level similarities, there are several essential differences between the DRM and the MFA-based models, all of which are critical for reproducing DCNs. 
First, in the DRM the only randomness is due to the choice of the $g^{\ell}$ and the observation noise after rendering. 
This naturally leads to inference of the most probable configuration via the max-sum algorithm, which is equivalent to max-pooling in the DCN. Second, the DRM's affine transformations $\Lambda^{\ell}$ act on {\em multi-channel} images at level $\ell+1$ to produce {\em multi-channel} images at level $\ell$. This structure is important, because it leads directly to the notion of (multi-channel) feature maps in DCNs. Third, a DRM's layers vary in connectivity from sparse to dense, as they give rise to convolutional, locally connected, and fully connected layers in the resulting DCN. Fourth, the DRM has switching variables that model (in)active renderers (Section~\ref{sec:rm}). The manifestation of these variables in the DCN are the ReLus (Eq.~\ref{eqn:relu}). Thus, the critical elements of the DCN architecture arise directly from aspects of the DRM structure that are absent in MFA-based models.

\subsection{\emph{i}-Theory: Invariant Representations Inspired by Sensory Cortex }

\emph{Representational Invariance and selectivity (RI)} are important ideas that have developed in the computational neuroscience community.
According to this perspective, the main purpose of the feedforward aspects of visual cortical processing in the ventral stream are to compute a representation for a sensed image that is invariant to irrelevant transformations (e.g., pose, lighting etc.) \cite{Pinto:2008bo,DiCarlo:2012em}. In this sense, the RI perspective is quite similar to the DRM in its basic motivations. However, the RI approach has remained qualitative in its explanatory power until recently, when a theory of invariant representations in deep architectures --- dubbed \emph{i-theory} ---  was proposed \cite{Anselmi:2007ke,PoggioOnInvariance}. Inspired by neuroscience and models of the visual cortex, it is the first serious attempt at explaining the success of deep architectures, formalizing intuitions about invariance and selectivity in a rigorous and quantitatively precise manner. 

The \ith posits a representation that employs group averages and orbits to explicitly insure invariance to specific types of nuisance transformations. These transformation must possess a mathematical semi-group structure; as a result, the invariance constraint is relaxed to a notion of partial invariance, which is built up slowly over multiple layers of the architecture.

At a high level, the DRM shares similar goals with \ith in that it attempts to capture explicitly the notion of nuisance transformations. However, the DRM differs from \ith in two critical ways. First, it does not impose a semi-group structure on the set of nuisance transformations. This provides the DRM the flexibility to learn a representation that is invariant to a wider class of nuisance transformations, including non-rigid ones. Second, the DRM does not fix the representation for images in advance. Instead, the representation emerges naturally out of the inference process. For instance, sum- and max-pooling emerge as probabilistic marginalization over nuisance variables and thus are necessary for proper inference. The deep iterative nature of the DCN also arises as a direct mathematical consequence of the DRM's rendering model, which comprises multiple levels of abstraction.

This is the most important difference between the two theories. Despite these differences, \ith is complementary to our approach in several ways, one of which is that it spends a good deal of energy focusing on questions such as: How many templates are required for accurate discrimination? How many samples are needed for learning? We plan to pursue these questions for the DRM in future work.

\subsection{Scattering Transform: Achieving Invariance via Wavelets}

We have used the DRM, with its notion of target and nuisance variables, to explain the power of DCN for learning selectivity and invariance to nuisance transformations. Another theoretical approach to learning selectivity and invariance is the \emph{Scattering Transform} (ST) \cite{bruna2013invariant,mallat2012group}, which consists of a series of linear wavelet transforms interleaved by nonlinear modulus-pooling of the wavelet coefficients. The goal is to \emph{explicitly hand-design invariance} to a specific set of nuisance transformations (translations, rotations, scalings, and small deformations) by using the properties of wavelet transforms. 

If we ignore the modulus-pooling for a moment, then the ST implicitly assumes that images can be modeled as linear combinations of pre-determined wavelet templates. Thus the ST approach has a maximally strong model bias, in that there is no learning at all. The ST performs well on tasks that are consistent with its strong model bias, i.e., on small datasets for which successful performance is therefore contingent on strong model bias. However, the ST will be more challenged on difficult real-world tasks with complex nuisance structure for which large datasets are available. This contrasts strongly with the approach presented here and that of the machine learning community at large, where hand-designed features have been outperformed by learned features in the vast majority of tasks.

\subsection{Learning Deep Architectures via Sparsity}

What is the optimal machine learning architecture to use for a given task? This question has typically been answered by exhaustively searching over many different architectures. But is there a way to \emph{learn} the optimal architecture directly from the data?
Arora et al.\ \cite{arora2013provable} provide some of the first theoretical results in this direction. In order to retain theoretical tractability, they assume a simple sparse neural network as the generative model for the data. Then, given the data, they design a greedy learning algorithm that reconstructs the architecture of the generating neural network, layer-by-layer. 

They prove that their algorithm is optimal under a certain set of restrictive assumptions. Indeed, as a consequence of these restrictions, their results do not directly apply to the DRM or other plausible generative models of natural images. However, the core message of the paper has nonetheless been influential in the development of the \emph{Inception} architecture \cite{szegedy2014going}, which has recently achieved the highest accuracy on the ImageNet classification benchmark \cite{ioffe2015batch}.

How does the sparse reconstruction approach relate to the DRM? The DRM is indeed also a sparse generative model: the act of rendering an image is approximated as a sequence of affine transformations applied to an abstract high-level class template. Thus, the DRM can potentially be represented as a sparse neural network. Another similarity between the two approaches is the focus on clustering  highly correlated activations in the next coarser layer of abstraction. Indeed the DRM is a composition of sparse factor analyzers, and so each higher layer $\ell+1$ in a DCN really does decorrelate and cluster the layer $\ell$ below, as quantified by Eq.~\ref{eqn:FwdProp}.

But despite these high-level similarities, the two approaches differ significantly in their overall goals and results. First, our focus has not been on recovering the architectural parameters; instead we have focused on the class of architectures that are well-suited to the task of factoring out large amounts of nuisance variation. In this sense the goals of the two approaches are different and complementary. Second, we are able to derive the structure of DCNs and RDFs exactly from the DRM. This enables us to bring to bear the full power of probabilistic analysis for solving high-nuisance problems; moreover, it will enable us to build better models and representations for hard tasks by addressing limitations of current approaches in a principled manner.

\subsection{Google FaceNet: Learning Useful Representations with DCNs}

Recently, Google developed a new face recognition architecture called \emph{FaceNet} \cite{schroff2015facenet} that illustrates the power of learning good representations.  It achieves state-of-the-art accuracy in face recognition and clustering on several public benchmarks.
FaceNet uses a DCN architecture, but crucially, it was not trained for classification. Instead, it is trained to optimize a novel learning objective called \emph{triplet finding} that learns good representations in general. 

The basic idea behind their new representation-based learning objective is to \emph{encourage the DCN's latent representation to embed images of the same class close to each other while embedding images of different classes far away from each other}, an idea that is similar to the NuMax algorithm\cite{hegde2012convex}. In other words, the learning objective enforces a \emph{well-separatedness} criterion. In light of our work connecting DRMs to DCNs, we will next show how this new learning objective can be understood from the perspective of the DRM.

The correspondence between the DRM and the triplet learning objective is simple. Since rendering is a deterministic (or nearly noise-free) function of the global configuration $(c,g)$, one explanation should dominate for any given input image $I = R(c,g)$, or equivalently, the clusters $(c,g)$ should be well-separated.
Thus, the noise-free, deterministic, and well-separated DRM are all equivalent. 
Indeed, we implicitly used the well-separatedness criterion when we employed the Hard EM algorithm to establish the correspondence between DRMs and DCNs/RDFs. 

\subsection{Renormalization Theory}

Given the DRM's notion of irrelevant (nuisance) transformations and multiple levels of abstraction, we can interpret a DCN's action as an {\em iterative coarse-graining} of an image, thus relating our work to another recent approach to understanding deep learning that draws upon an analogy from renormalization theory in physics\cite{mehta2014exact}.  This approach constructs an exact correspondence between the Restricted Boltzmann Machine (RBM) and block-spin renormalization --- an iterative coarse-graining technique from physics that compresses a configuration of binary random variables (spins) to a smaller configuration with less variables. The goal is to preserve as much information about the longer-range correlations as possible, while integrating out shorter-range fluctuations. 

Our work here shows that this analogy goes even further as we have created an exact mapping between the DCN and the DRM, the latter of which can be interpreted as a new real-space renormalization scheme. 
Indeed, the DRM's main goal is to factor out irrelevant features over multiple levels of detail, and it thus bears a strong resemblance to the core tenets of renormalization theory. As a result, we believe this will be an important avenue for further research. 

\subsection{Summary of Key Distinguishing Features of the DRM}

The key features that distinguish the DRM approach from others in the literature can be summarized as:
(i) The DRM explicitly models nuisance variation across multiple levels of abstraction via a product of affine transformations. This factorized linear structure serves dual purposes: it enables (ii) exact inference (via the max-sum/max-product algorithm) and (iii) it serves as a regularizer, preventing overfitting by a novel exponential reduction in the number of parameters. 
Critically, (iv) the inference is not performed for a single variable of interest but instead for the full global configuration. This is justified in low-noise settings, i.e., when the rendering process is nearly deterministic, and suggests the intriguing possibility that vision is less about probabilities and more about inverting a complicated (but deterministic) rendering transformation.

\section{New Directions} 
\label{sec:ext}

We have shown that the DRM is a powerful generative model that underlies both DCNs and RDFs, the two most powerful vision paradigms currently employed in machine learning. 
Despite the power of the DRM/DCN/RDF, it has limitations, and there is room for improvement. 
(Since both DCNs and RDFs stem from DRMs, we will loosely refer to them both as DCNs in the following, although technically an RDF corresponds to a kind of tree of DCNs.)

In broad terms, most of the limitations of the DCN framework can be traced back to the fact that it is a discriminative classifier whose underlying generative model was not known. 
Without a generative model, many important tasks are very difficult or impossible, including sampling, model refinement, top-down inference, faster learning, model selection, and learning from unlabeled data. 
With a generative model, these tasks become feasible. 
Moreover, the DCN models rendering as a sequence of affine transformations, which severely limits its ability to capture many important real-world visual phenomena, including figure-ground segmentation, occlusion/clutter, and refraction. 
It also lacks several operations that appear to be fundamental in the brain: feed-back, dynamics, and 3D geometry. 
Finally, it is unable to learn from unlabeled data and to generalize from few examples. As a result, DCNs require enormous amounts of labeled data for training. 

These limitations can be overcome by designing new deep networks based on new model structures (extended DRMs), new message-passing inference algorithms, and new learning rules, as summarized in Table~\ref{fig:LimitationsOfDCNs}. We now explore these solutions in more detail.

\begin{table}
   \centering
      \includegraphics[width=0.80\linewidth]{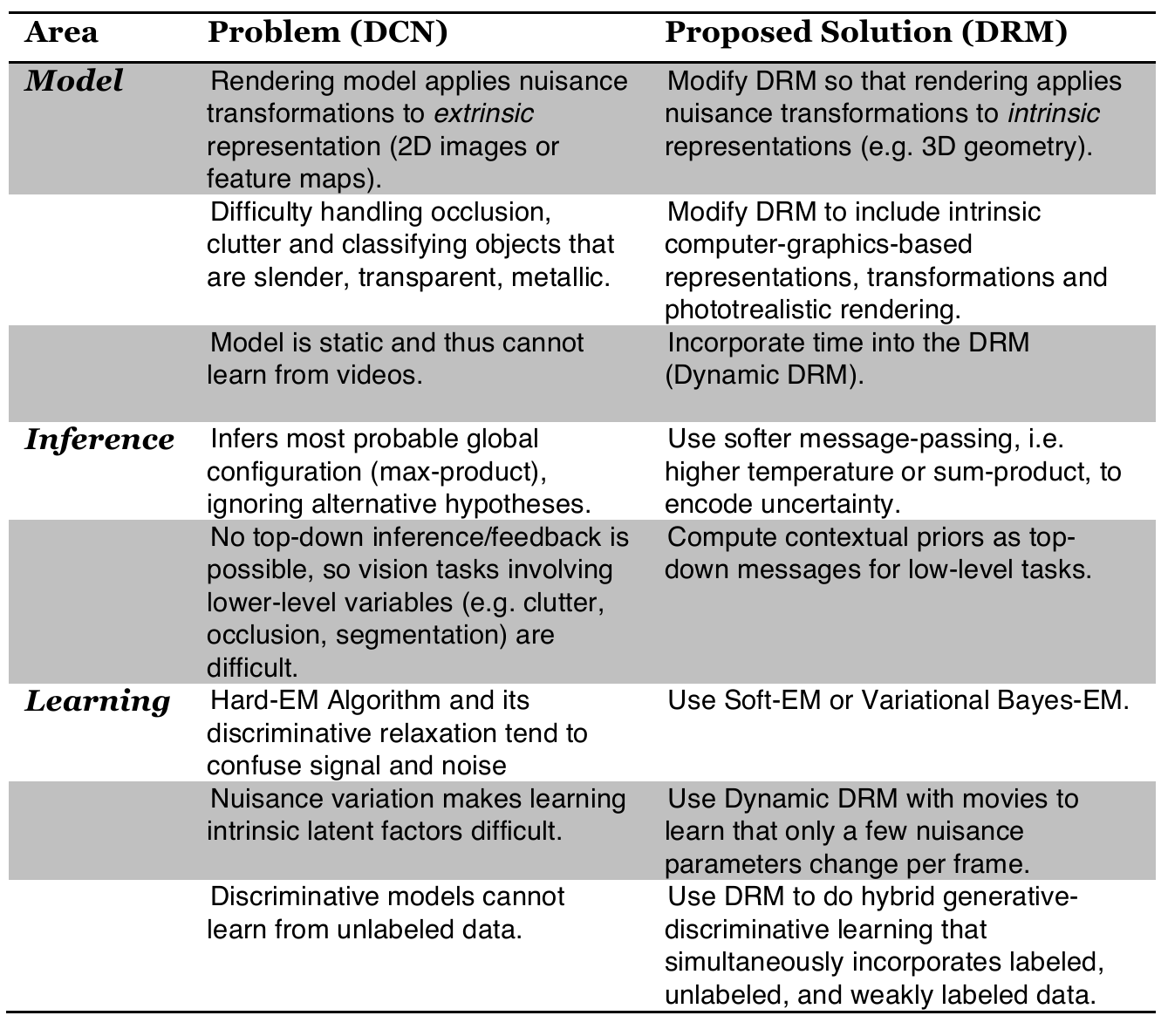} 
   \caption{Limitations of current DCNs and potential solutions using extended DRMs.}
   \label{fig:LimitationsOfDCNs}
\end{table}

\subsection{More Realistic Rendering Models}

We can improve DCNs by designing better generative models incorporating more realistic assumptions about the rendering process by which latent variables cause images. These assumptions should include symmetries of translation, rotation, scaling \cite{bruna2013invariant}, perspective, and non-rigid deformations, as rendered by computer graphics and multi-view geometry. 

In order to encourage more intrinsic computer graphics-based representations, we can enforce these symmetries on the parameters during learning \cite{miao2007learning, anselmi2013unsupervised}. Initially, we could use local affine approximations to these transformations \cite{sohl2010unsupervised}. For example, we could impose weight tying based on 3D rotations in depth. Other nuisance transformations are also of interest, such as scaling (i.e., motion towards or away from a camera).
Indeed, scaling-based templates are already in use by the state-of-the-art DCNs such as the \emph{Inception} architectures developed by Google \cite{szegedy2014going}, and so this approach has already shown substantial promise.  

We can also perform intrinsic transformations directly on 3D scene representations. For example, we could train networks with depth maps, in which a subset of channels in input feature maps encode pixel $z$-depth. These augmented input features will help define useful higher-level features for 2D image features, and thereby transfer representational benefits even to test images that do not provide depth information \cite{hartley2003multiple}. With these richer geometric representations, learning and inference algorithms can be modified to account for 3D constraints according to the equations of multi-view geometry \cite{hartley2003multiple}.

Another important limitation of the DCN is its restriction to static images. There is no notion of time or dynamics in the corresponding DRM model. As a result, DCN training on large-scale datasets requires millions of images in order to learn the structure of high-dimensional nuisance variables, resulting in a glacial learning process. In contrast, learning from natural videos should result in an accelerated learning process, \emph{as typically only a few nuisance variables change from frame to frame}. This property should enable substantial acceleration in learning, as inference about which nuisance variables have changed will be faster and more accurate \cite{michalski2014modeling}.  See Section~\ref{sec:learning-dyn} below for more details.

\subsection{New Inference Algorithms}

\subsubsection{Soft Inference} 

We showed above in Section \ref{sec:infDRM} that DCNs implicitly infer the most probable global interpretation of the scene, via the max-sum algorithm \cite{pearl1988probabilistic}. However, there is potentially major component missing in this algorithm: max-sum message passing only propagates the most likely hypothesis to higher levels of abstraction, which may not be the optimal strategy, in general, especially if uncertainty in the measurements is high (e.g., vision in a fog or at nighttime). Consequently, we can consider a wider variety of \emph{softer} inference algorithms by defining a temperature parameter that enables us to smoothly interpolate between the max-sum and sum-product algorithms, as well as other message-passing variants such as the approximate Variational Bayes EM \cite{bishop2006pattern}. To the best of our knowledge, this notion of a soft DCN is novel.

\subsubsection{Top-Down Convolutional Nets: Top-Down Inference via the DRM} \label{sec:top-down-dcn} 

The DCN inference algorithm lacks any form of top-down inference or feedback. Performance on tasks using low-level features is then suboptimal, because higher-level information informs low-level variables neither for inference nor for learning. We can solve this problem by using the DRM, since it is a proper generative model and thus enables us to implement top-down message passing properly.  

Employing the same steps as outlined in Section~\ref{sec:drm}, we can convert the DRM into a {\em top-down DCN}, a neural network that implements both the bottom-up and top-down passes of inference via the max-sum message passing algorithm. This kind of top-down inference should have a dramatic impact on scene understanding tasks that require segmentation such as target detection with occlusion and clutter, where local bottom-up hypotheses about features are ambiguous. To the best of our knowledge, this is the first principled approach to defining top-down DCNs.

\subsection{New Learning Algorithms}

\subsubsection{Derivative-Free Learning}

Back propagation is often used in deep learning algorithms due to its simplicity. We have shown above that back propagation in DCNs is actually an inefficient implementation of an approximate EM algorithm, whose E-step consists of bottom-up inference and whose M-step is a gradient descent step that fails to take advantage of the underlying probabilistic model (the DRM). 
To the contrary, our above EM algorithm (Eqs.~\ref{eqn:FwdProp}--\ref{eqn:MDRM}) is both much faster and more accurate, because it directly exploits the DRM's structure. Its E-step incorporates bottom-up and top-down inference, and its M-step is a fast computation of sufficient statistics (e.g., sample counts, means, and covariances). The speed-up in efficiency should be substantial, since generative learning is typically much faster than discriminative learning due to the bias-variance tradeoff \cite{jordan2002discriminative}; moreover, the EM-algorithm is intrinsically more parallelizable \cite{kumar2009fast}. 

\subsubsection{Dynamics: Learning from Video} 
\label{sec:learning-dyn}

Although deep NNs have incorporated time and dynamics for auditory tasks \cite{hochreiter1997long, graves2013speech, graves2013hybrid}, DCNs for visual tasks have remained predominantly static (images as opposed to videos) and are trained on static inputs. Latent causes in the natural world tend to change little from frame-to-frame, such that previous frames serve as partial self-supervision during learning \cite{wiskott2006does}. 
A dynamic version of the DRM 
would train without external supervision on large quantities of video data (using the corresponding EM algorithm). 
We can supplement video recordings of natural dynamic scenes with synthetically rendered videos of objects traveling along smooth trajectories, which will enable the training to focus on learning key nuisance factors that cause difficulty (e.g., occlusion).

\subsubsection{Training from Labeled and Unlabeled Data}

DCNs are purely discriminative techniques and thus cannot benefit from unlabeled data. However, armed with a generative model we can perform \emph{hybrid discriminative-generative training \cite{bishop2007generative}} that enables training to benefit from both labeled and unlabeled data in a principled manner. This should dramatically increase the power of pre-training, by encouraging representations of the input that have disentangled factors of variation. This hybrid generative-discriminative learning is achieved by the optimization of a novel objective function for learning, that relies on both the generative model \emph{and} its discriminative relaxation. In particular, the learning objective will have terms for both, as described in \cite{bishop2007generative}. Recall from Section~\ref{sec:gen-to-discr}  that the discriminative relaxation of a generative model is performed by relaxing certain parameter constraints during learning, according to
\begin{align} 
	\max_{\theta} L_{\rm gen}(\theta; \mathcal{D}_{CI}) &= \max_{\eta : \eta = \rho(\theta)} L_{\rm nat}(\eta; \mathcal{D}_{CI}) \nonumber\\
	&\leq \max_{\eta : \eta = \rho(\theta)} L_{\rm cond}(\eta; \mathcal{D}_{C|I}) \nonumber\\
	&\leq \max_{\eta}L_{\rm dis}(\eta; \mathcal{D}_{C|I}),
\end{align}
where the $L$'s are the model's generative, naturally parametrized generative, conditional, and discriminative likelihoods. Here $\eta$ are the natural parameters expressed as a function of the traditional parameters $\theta$, $\mathcal{D}_{CI}$ is the training dataset of labels and images, and $\mathcal{D}_{C|I}$ is the training dataset of labels given images. Although the discriminative relaxation is optional, it is very important for achieving high performance in real-world classifiers as discriminative models have less model bias and, therefore, are less sensitive to model mis-specifications \cite{jordan2002discriminative}. Thus, we will design new principled training algorithms that span the spectrum from discriminative (e.g., Stochastic Gradient Descent with Back Propagation) to generative (e.g., EM Algorithm).

\section*{Acknowledgments}
Thanks to CJ Barberan for help with the manuscript and to Mayank Kumar, Ali Mousavi, Salman Asif and Andreas Tolias for comments and discussions. Thanks to Karen Simonyan for providing the activity maximization figure. A special thanks to Xaq Pitkow whose keen insight, criticisms and detailed feedback on this work have been instrumental in its development. Thanks to Ruchi Kukreja for her unwavering support and her humor and to Raina Patel for providing inspiration.

\clearpage

\clearpage
\appendix

\section{Supplemental Information}
\label{sec:supp}

\subsection{From the Gaussian Rendering Model Classifier to Deep DCNs }

\begin{proposition}[MaxOut NNs] \label{eq:GRM-MaxOutNN}
The discriminative relaxation of a noise-free GRM classifier is a single layer NN consisting of a local template matching operation followed by a piecewise linear activation function (also known as a \emph{MaxOut NN}\cite{goodfellow2013maxout}).
\end{proposition}
\begin{proof}
In order to teach the reader, we prove this claim exhaustively. Later claims will have simple proofs that exploit the fact that the RM's distribution is from the exponential family.

\begin{align*}
	\hat{c}(I)                 &\equiv \argmax_{c \in \Cl} p(c|I) \\
	                               &= \argmax_{c \in \Cl} \left\{ p(I | c)  p(c) \right\} \\
	                               &= \argmax_{c \in \Cl} \left\{ \sum_{h \in \HH} p(I | c,h)  p(c,h) \right\} \\
	                               &\overset{(a)}{=} \argmax_{c \in \Cl} \left\{ \max_{h \in \HH} p(I | c,h)  p(c,h) \right\} \\
	                               &= \argmax_{c \in \Cl} \left\{ \max_{h \in \HH} \exp \left( \ln p(I | c,h)  + \ln p(c,h) \right) \right\} \\
	                               &\overset{(b)}{=} \argmax_{c \in \Cl} \left\{ \max_{h \in \HH} \exp \left( \sum_{\omega} \ln p(I^{\omega} | c,h)  + \ln p(c,h) \right) \right\} \\
	                               &\overset{(c)}{=} \argmax_{c \in \Cl} \left\{ \max_{h \in \HH} \exp \left( -\frac{1}{2} \sum_{\omega} \left< I^{\omega} - \mu_{ch}^{\omega} | \Sigma^{-1}_{ch} | I^{\omega} - \mu_{ch}^{\omega} \right>  + \ln p(c,h) -\frac{D}{2} \ln |\Sigma_{ch}| \right) \right\} \\
	                               &= \argmax_{c \in \Cl} \left\{ \max_{h \in \HH} \exp \left( \sum_{\omega} \left< w_{ch}^{\omega} | I^{\omega} \right>  + b_{ch}^{\omega} \right) \right\} \\
	                               &\overset{(d)}{\equiv} \argmax_{c \in \Cl} \left\{ \exp \left( \max_{h \in \HH}  \left\{ w_{ch} \lcmatch I \right\}  \right) \right\} \\
	                               &= \argmax_{c \in \Cl} \left\{ \max_{h \in \HH}  \left\{ w_{ch} \lcmatch I \right\} \right\} \\	                               
	                               &= \textrm{Choose} \left\{ \textrm{MaxOutPool}(  \textrm{LocalTemplateMatch} (I))  \right\} \\
	                               &= \textrm{MaxOut-NN}(I; \theta).
\end{align*}
In line (a), we take the noise-free limit of the GRM, which means that one hypothesis $(c,h)$ dominates all others in likelihood. In line (b), we assume that the image $I$ consists of multiple channels $\omega \in \Omega$, that are conditionally independent given the global configuration $(c,h)$. Typically, for input images these are color channels and $\Omega \equiv \{ r,g,b \}$ but in general $\Omega$ can be more abstract (e.g. as in feature maps). In line (c), we assume that the pixel noise covariance is isotropic and conditionally independent given the global configuration $(c,h)$, so that $\Sigma_{ch} = \sigma_{x}^{2} \mathbf{1}_{D} $ is proportional to the $D \times D$ identity matrix $\mathbf{1}_{D}$. In line (d), we defined the \emph{locally connected template matching operator} $\lcmatch$, which is a location-dependent template matching operation.
\end{proof}

Note that the nuisance variables $h \in \HH$ are (max-)marginalized over, after the application of a local template matching operation against a set of filters/templates $\W \equiv \{ w_{ch} \}_{c \in \Cl, h \in \HH}$

\begin{lemma}[\textbf{Translational Nuisance $\drelax$ DCN Convolution}] 
\label{lem:trans-to-dcn-conv}
 The MaxOut template matching and pooling operation (from Proposition \ref{eq:GRM-MaxOutNN}) for a set of translational nuisance variables $\HH \equiv \G_{T}$ reduces to the traditional DCN convolution and max-pooling operation.
\end{lemma}

\begin{proof}
 Let the activation for a single output unit be $y_{c}(I)$. Then we have
 \begin{align*} 
    y_{c}(I) &\equiv  \max_{h \in \HH}  \left\{ w_{ch} \lcmatch I \right\}   \\
                &=  \max_{g \in \G_{T}}  \left\{ \left< w_{cg} | I \right> \right\}  \\
                &=  \max_{g \in \G_{T}}  \left\{ \left< T_{g} w_{c} | I \right> \right\}  \\
                &=  \max_{g \in \G_{T}}  \left\{ \left< w_{c} | T_{-g} I \right> \right\}  \\
                &=  \max_{g \in \G_{T}}  \left\{ (w_{c} \cnnconv I)_{g} \right\}  \\
                &=  \MaxPool( w_{c} \cnnconv I).
 \end{align*}
 Finally, vectorizing in $c$ gives us the desired result $y(I) = \MaxPool( \W \cnnconv I)$.
\end{proof}

\begin{proposition}[Max Pooling DCNs with ReLu Activations] \label{prop:detailedReLuProof}
The discriminative relaxation of a noise-free GRM \textbf{with translational nuisances and random missing data} is a single convolutional layer of a traditional DCN. The layer consists of a generalized convolution operation, followed by a ReLu activation function and a Max-Pooling operation.
\end{proposition}
\begin{proof}
We will model completely random missing data as a nuisance transformation $a \in \A \equiv \{\textrm{keep}, \textrm{drop} \}$, where $a=\textrm{keep}=1$ leaves the rendered image data untouched, while $a=\textrm{drop}=0$ throws out the entire image after rendering. Thus, the switching variable $a$ models missing data. Critically, whether the data is missing is assumed to be \textit{completely random}  and thus independent of any other task variables, including the measurements (i.e. the image itself). Since the missingness of the evidence is just another nuisance, we can invoke Proposition \ref{eq:GRM-MaxOutNN} to conclude that the discriminative relaxation of a noise-free GRM with random missing data is also a MaxOut-DCN, but with a specialized structure which we now derive.

Mathematically, we decompose the nuisance variable $h \in \HH$ into two parts $h = (g,a) \in \HH = \G \times \A$, and then, following a similar line of reasoning as in Proposition \ref{eq:GRM-MaxOutNN}, we have
\begin{align*}
	\hat{c}(I)                 &= \argmax_{c \in \Cl} \max_{h \in \HH} p(c,h|I) \\
	                               &= \argmax_{c \in \Cl} \left\{ \max_{h \in \HH}  \left\{ w_{ch} \lcmatch I \right\} \right\} \\
	                               &\overset{(a)}{=} \argmax_{c \in \Cl} \left\{ \max_{g \in \G} \max_{a \in \A}  \left\{ a(\langle w_{cg} | I\rangle + b_{cg})   + b_{cg}' + b_{a} + b_{I}'\right\} \right\} \\
	                               &\overset{(b)}{=} \argmax_{c \in \Cl} \left\{ \max_{g \in \G}  \left\{ \max \{ (w_{c} \cnnconv I)_{g}, 0 \right\} + b_{cg}' + b_{\textrm{drop}}' + b_{I}' \} \right\} \\
	                               &\overset{(c)}{=} \argmax_{c \in \Cl} \left\{ \max_{g \in \G}  \left\{ \max \{ (w_{c} \cnnconv I)_{g}, 0 \right\} + b_{cg}' \} \right\} \\
	                               &\overset{(d)}{=} \argmax_{c \in \Cl} \left\{ \max_{g \in \G}  \left\{ \max \{ (w_{c} \cnnconv I)_{g}, 0 \right\} \} \right\} \\
	                               &= \textrm{Choose} \left\{ \textrm{MaxPool}( \textrm{ReLu} ( \textrm{DCNConv} (I)))  \right\} \\
	                               &= \textrm{DCN}(I; \theta).
\end{align*}
In line (a) we calculated the log-posterior
\begin{align*}
	\ln p(c,h | I)  &= \ln p(c,g,a | I)  \\
			&= \ln p(I | c,g,a)  + \ln p(c,g,a) \\
			&= \frac{1}{2\sigma_x^2} \langle a \mu_{cg} | I \rangle  -\frac{1}{2\sigma_x^2} (\| a \mu_{cg} \|_2^{2} + \| I \|_2^{2})  )  + \ln p(c,g,a)\\
	                 &\equiv  a(\langle w_{cg} | I\rangle + b_{cg}) +  b_{cg}' + b_{a} + b_{I}',
\end{align*}
where $a \in \{0,1\}, b_{a} \equiv \ln p(a), b_{cg}' \equiv \ln p(c,g), b_{I}' \equiv -\frac{1}{2\sigma_x^2} \| I \|_{2}^{2}$. In line (b), we use Lemma \ref{lem:trans-to-dcn-conv} to write the expression in terms of the DCN convolution operator, after which we invoke the identity $\max \{u,v\} = \max \{u-v,0 \}+v \equiv \textrm{ReLu}(u-v) + v$ for real numbers $u,v \in \R$. Here we've defined $b_{\textrm{drop}}' \equiv \ln p(a=\textrm{keep})$ and we've used a slightly modified DCN convolution operator $\cnnconv$ defined by $w_{cg} \cnnconv I \equiv w_{cg} \star I + \ln \left( \frac{p(a=\textrm{keep})}{p(a=\textrm{drop})} \right)$. Also, we observe that all the primed constants are independent of $a$ and so can be pulled outside of the $\max_{a}$. In line(c), the two primed constants that are also independent of $c,g$ can be dropped due to the $\argmax_{cg}$. Finally, in line (d), we assume a uniform prior over $c,g$.
The resulting sequence of operations corresponds \emph{exactly} to those applied in a single convolutional layer of a traditional DCN.
\end{proof}

\begin{remark}[\textbf{The Probabilistic Origin of the Rectified Linear Unit}] \sloppy
Note the origin of the ReLu in the proof above: it compares the relative (log-)likelihood of two hypotheses $a=\textrm{keep}$ and $a=\textrm{drop}$, i.e. whether the current measurements (image data $I$) are available/relevant/important or instead missing/irrelevant/unimportant for hypothesis $(c,g)$. In this way, the ReLu also promotes sparsity in the activations.
\end{remark}

\subsection{Generalizing to Arbitrary Mixtures of Exponential Family \\ Distributions} \label{sec:generalize-exp-fam}

In the last section, we showed that the GRM -- a mixture of Gaussian Nuisance Classifiers -- has as its discriminative relaxation a MaxOut NN. In this section, we generalize this result to an arbitrary mixture of Exponential family Nuisance classifiers. For example, consider a Laplacian RM (LRM) or a Poisson RM (PRM). 

\begin{definition}[Exponential Family Distributions] \label{defn:ExpFamily}
 A distribution $p(x;\theta)$ is in the exponential family if it can be written in the form
 \[
 	p(x;\theta) = h(x) \exp ( \langle \eta(\theta) | T(x) \rangle - A(\eta)),
 \]
 where $\eta(\theta)$ is the vector of \textbf{natural parameters}, $T(x)$is the vector of \textbf{sufficient statistics}, $A(\eta(\theta))$ is the \textbf{log-partition function}.
\end{definition}

By generalizing to the exponential family, we will see that derivations of the discriminative relations will simplify greatly, with the key roles being played by familiar concepts such as natural parameters, sufficient statistics and log-partition functions. Furthermore, most importantly, we will see that the resulting discriminative counter parts are \emph{still} MaxOut NNs. Thus MaxOut NNs are quite a robust class, as most E-family mixtures have MaxOut NNs as d-counterparts.

\begin{theorem}[Discriminative Counterparts to Exponential Family Mixtures are MaxOut Neural Nets] \label{thm:DiscrExpFam}
 Let $\M_{g}$ be a Nuisance Mixture Classifier from the Exponential Family. Then the discriminative counterpart $\M_{d}$ of $\M_{g}$ is a MaxOut NN.
\end{theorem}

\begin{proof}
The proof is analogous to the proof of Proposition \ref{eq:GRM-MaxOutNN}, except we generalize by using the definition of an exponential family distribution (above). We simply use the fact that all exponential family distributions have a natural or canonical form as described above in the Definition~\ref{defn:ExpFamily}. Thus the natural parameters will serve as generalized weights and biases, while the sufficient statistic serves as the generalized input. Note that this may require a non-linear transformation i.e. quadratic or logarithmic, depending on the specific exponential family.
\end{proof}

\subsection{Regularization Schemes: Deriving the DropOut Algorithm}

Despite the large amount of labeled data available in many real-world vision applications of deep DCNs, regularization schemes are still a critical part of training, essential for avoiding overfitting the data. The most important such scheme is DropOut \cite{hinton2012improving} and it consist of training with unreliable neurons and synapses. Unreliability is modeled by a `dropout' probability $p_{d}$ that the neuron will not fire (i.e. output activation is zero) or that the synapse won't send its output to the receiving neuron. Intuitively, downstream neurons cannot rely on every piece of data/evidence always being there, and thus are forced to develop a robust set of features. This prevents the co-adaptation of feature detectors that undermines generalization ability.

In this section, we answer the question: Can we derive the DropOut algorithm from the generative modeling perspective? Here we show that the answer is yes. Dropout can be derived from the GRM generative model via the use of the EM algorithm under the condition of (completely random) missing data.

\begin{proposition} \label{prop:GRM-DropOut}
The discriminative relaxation of a noise-free GRM with completely random missing data is a DropOut DCN\cite{dahl2013improving} with Max-Pooling.
\end{proposition}
\begin{proof}
Since we have data that is \emph{missing completely at random}, we can use the EM algorithm to train the GRM\cite{bishop2006pattern}. Our strategy is to show that a single iteration of the EM-algorithm corresponds to a full epoch of DropOut DCN training (i.e. one pass thru the entire dataset). Note that typically an EM-algorithm is used to train generative models; here we utilize the EM-algorithm in a novel way, performing a discriminative relaxation in the M-step. In this way, we use the generative EM algorithm to define a \emph{discriminative EM algorithm (d-EM)}.

The d-E-step is equivalent to usual generative E-step. Given the observed data $X$ and the current parameter estimate $\hat{\theta}^{t}$, we will compute the posterior of the latent variables $Z = (H,A)$ where $A$ is the missing data indicator matrix i.e. $A_{np} = 1$ iff the $p$-th feature (e.g. pixel intensity) of the input data $I_{n}$ (e.g. natural image) is available. $H$ contains all other latent nuisance variables (e.g. pose) that are important for the classification task. Since we assume a noise-free GRM, we will actually execute a \emph{hybrid} E-step: hard in $H$ and soft in $A$. The hard-E step will yield the Max-Sum Message Passing algorithm, while the soft E-step will yield the ensemble average that is the characteristic feature of Dropout\cite{dahl2013improving}.

\sloppy
In the d-M-step, we will start out by maximizing the complete-data log-likelihood $\ell(\theta; H,A,X)$, just as in the usual generative M-step. However, near the end of the derivation \emph{we will employ a discriminative relaxation} that will free us from the rigid distributional assumptions of the generative model  $\theta_{g}$ and instead leave us with a much more flexible set of assumptions, as embodied in the discriminative modeling problem for $\theta_{d}$.

Mathematically, we have a single E-step and M-step that leads to a parameter update as follows:
\begin{align*} 
   \ell(\hat{\theta}_{\textrm{new}}) \equiv& \max_{\theta} \Big\{ \Expect_{Z|X}[ \ell(\theta;Z,X) ]   \Big\}     \\
                     =& \max_{\theta} \Big\{ \Expect_{A} \Expect_{H|X}[ \ell(\theta;H,A,X) ]   \Big\}     \\
                     =& \max_{\theta} \Big\{ \Expect_{A} \Expect_{H|X}\Big[ \ell(\theta;C,H | I,A) + \ell(\theta;I) + \ell(\theta;A) \Big]   \Big\}     \\
                     =& \max_{\theta_{d} \sim_{d} \theta_{g}} \Big\{ \Expect_{A} \Expect_{H|X}\Big[ \ell(\theta_{d};C,H | I,A) + \ell(\theta_{g};I) \Big]   \Big\}     \\
                     \leq& \max_{\theta_{d}} \Big\{ \Expect_{A} \Expect_{H|X}\Big[ \ell(\theta_{d};C,H | I,A) \Big]   \Big\}     \\
                     =& \max_{\theta_{d}} \Big\{ \Expect_{A} \MaxExp_{H|X}\Big[ \ell(\theta_{d};C,H | I,A) \Big]   \Big\}     \\
                     \equiv& \max_{\theta_{d}} \Big\{ \Expect_{A} \Big[ \ell(\theta_{d};C,H^{*} | I,A) \Big]   \Big\}     \\
                     =& \max_{\theta_{d}} \Big\{ \sum_{A} p(A) \cdot \ell(\theta_{d};C,H^{*} | I,A)  \Big\}     \\
                     \approx& \max_{\theta_{d}} \Big\{ \sum_{A\in \mathcal{T}} p(A) \cdot \ell(\theta_{d};C,H^{*} | I,A)  \Big\}     \\
                     =& \max_{\theta_{d}} \Big\{ \sum_{A \in \mathcal{T}} p(A) \cdot \sum_{n \in \mathcal{D_{CI}^{\textrm{dropout}}}}\ln p(c_{n}, h_{n}^{*} | I_{n}^{\textrm{dropout}}; \theta_{d})  \Big\} .
\end{align*}
Here we have defined the \emph{conditional likelihood} $\ell(\theta; D_{1} | D_{2}) \equiv \ln p(D_{1} | D_{2}; \theta)$, and $D= (D_{1}, D_{2})$ is some partition of the data. This definition allows us to write $\ell(\theta; D) = \ell(\theta; D_{1} | D_{2}) + \ell(\theta; D_{2})$ by invoking the conditional probability law $p(D | \theta) = p(D_{1} | D_{2} ; \theta) \cdot p(D_{2} | \theta)$. The symbol $\MaxExp_{H|X}[f(H)] \equiv \max_{H} \{ p(H|X) f(H) \}$ and the reduced dataset $\mathcal{D_{CI}^{\textrm{dropout}}}(A)$ is simply the original dataset of labels and features less the missing data (as specified by $A$).

The final objective function left for us to optimize is a mixture of exponentially-many discriminative models, each trained on a different random subset of the training data, but all sharing parameters (weights and biases). Since the sum over $A$ is intractable, we approximate the sums by Monte Carlo sampling of $A$ (the soft part of the E-step), yielding an ensemble $\mathcal{E} \equiv \{ A^{(i)} \}$. The resulting optimization corresponds exactly to the DropOut algorithm.
\end{proof}

\clearpage

\bibliography{MyReferences}
\bibliographystyle{IEEEtran}

\end{document}